\documentclass[sigconf]{acmart}

\usepackage{inputenc} 
\usepackage[T1]{fontenc}    
\usepackage{hyperref}       
\usepackage{url}            
\usepackage{booktabs}       
\usepackage{amsfonts}       
\usepackage{nicefrac}       
\usepackage{microtype}      
\usepackage{float}
\usepackage{algorithm}
\usepackage{makecell}
\usepackage{multirow}
\usepackage{hhline}
\usepackage{algorithmic}
\usepackage{amsmath,bm}
\usepackage[shortlabels]{enumitem}
\usepackage{mathrsfs}
\usepackage{graphicx}
\usepackage{color}
\usepackage{soul}
\usepackage{wrapfig}
\usepackage{subfigure}

\renewcommand{\a}{\mathbf{a}}

\renewcommand{\b}{\mathbf{b}}

\newcommand{\bg}{\bar{\mathbf{g}}}

\newcommand{\D}{\mathcal{D}}

\newcommand{\Eb}{\mathbb{E}}

\newcommand{\g}{\mathbf{g}}

\newcommand{\I}{\mathbf{I}}

\renewcommand{\u}{\mathbf{u}}

\renewcommand{\v}{\mathbf{v}}
\newcommand{\bv}{\bar{\mathbf{v}}}
\newcommand{\W}{\mathbf{W}}
\newcommand{\w}{\mathbf{w}}

\newcommand{\X}{\mathbf{X}}

\newcommand{\x}{\mathbf{x}}

\newcommand{\y}{\mathbf{y}}

\newcommand{\xb}{\mathbf{\bar{x}}}

\newcommand{\1}{\mathbf{1}}

\newtheorem{thm}{Theorem}

\newtheorem{cor}[thm]{Corollary}
\newtheorem{lem}{Lemma}

\newtheorem{defn}{Definition}

\newtheorem{assum}{Assumption}

\AtBeginDocument{%
  \providecommand\BibTeX{{%
    \normalfont B\kern-0.5em{\scshape i\kern-0.25em b}\kern-0.8em\TeX}}}

\copyrightyear{2020}
\acmYear{2020}
\setcopyright{rightsretained}
\acmDOI{10.1145/1122445.1122456}

\acmConference[MobiHoc '20]{MobiHoc '21: ACM International Symposium on Theory, Algorithmic Foundations, and Protocol Design for Mobile Networks and Mobile Computing}{June 30 -- July 03, 2020}{Shanghai, China}
\acmBooktitle{MobiHoc '20: ACM Symposium on Theory, Algorithmic Foundations, and Protocol Design for Mobile Networks and Mobile Computing,
  June 30 -- July 03, 2020, Shanghai, China}
\acmPrice{15.00}
\acmISBN{978-1-4503-XXXX-X/18/06}

\begin{document}

\title[{\fontsize{6pt}{7pt}\selectfont{GT-STORM: Taming Sample, Communication, and Memory Complexities in Decentralized Non-Convex Learning}}]{GT-STORM: Taming Sample, Communication, and Memory Complexities in Decentralized Non-Convex Learning}

\author{Xin Zhang$^1$, Jia Liu$^2$, Zhengyuan Zhu$^1$, and Elizabeth S. Bentley$^3$}
\affiliation{
\institution{$^1$Department of Statistics, Iowa State University}
\institution{$^2$Department of Electrical and Computer Engineering, The Ohio State University}
\institution{$^3$Information Directorate, Air Force Research Laboratory}
}

%
%
%
%

\renewcommand{\shortauthors}{Xin Zhang, Jia Liu, Zhengyuan Zhu, and Elizabeth Serena Bentley}

\begin{abstract}
Decentralized nonconvex optimization has received increasing attention in recent years in machine learning due to its advantages in system robustness, data privacy, and implementation simplicity.
However, three fundamental challenges in designing decentralized optimization algorithms are how to reduce their sample, communication, and memory complexities.
In this paper, we propose a \underline{g}radient-\underline{t}racking-based \underline{sto}chastic \underline{r}ecursive
\underline{m}omentum (GT-STORM) algorithm for efficiently solving nonconvex optimization problems.
We show that to reach an $\epsilon^2$-stationary solution, the total number of sample evaluations of our algorithm is $\tilde{O}(m^{1/2}\epsilon^{-3})$ and the number of communication rounds is $\tilde{O}(m^{-1/2}\epsilon^{-3})$, which improve the $O(\epsilon^{-4})$ costs of sample evaluations and communications for the existing decentralized stochastic gradient algorithms.
We conduct extensive experiments with a variety of learning models, including non-convex logistical regression and convolutional neural networks, to verify our theoretical findings. 
Collectively, our results contribute to the state of the art of theories and algorithms for decentralized network optimization.
\end{abstract}




\maketitle

\section{Introduction}\label{sec:Intro}

In recent years, machine learning has witnessed enormous success in many areas, including image processing, natural language processing, online recommender systems, just to name a few.
From a mathematical perspective, training machine learning models amounts to solving an optimization problem.
However, with the rapidly increasing dataset sizes and the high dimensionality and the non-convex hardness of the training problem (e.g., due to the use of deep neural networks), training large-scale machine learning models by a single centralized machine has become inefficient and unscalable.
To address the efficiency and scalability challenges, an effective approach is to leverage {\em decentralized} computational resources in a computing network, which could follow a parameter server (PS)-worker architecture \cite{recht2011hogwild,zinkevich2010parallelized,dean2012large} or fully decentralized peer-to-peer network structure \cite{nedic2009distributed,lian2017can}.
Also, thanks to the robustness to single-point-of-failure, data privacy, and implementation simplicity, decentralized learning over computing networks has attracted increasing interest recently, and has been applied in various science and engineering areas (including dictionary learning \cite{chen2014dictionary}, multi-agent systems \cite{cao2012overview,zhou2011multirobot}, multi-task learning \cite{wang2018distributed,zhang2019distributed}, information retrieval \cite{ali2004tivo}, energy allocation \cite{jiang2018consensus}, etc.).

In the fast growing literature of decentralized learning over networks, a classical approach is the so-called network consensus optimization, which traces its roots to the seminal work by Tsitsiklis in 1984~\cite{tsitsiklis1984problems}.
Recently, network consensus optimization has gained a lot of renewed interest owing to the elegant decentralized subgradient descent method (DSGD) proposed by Nedic and Ozdaglar \cite{nedic2009distributed}, which has been applied in decentralized learning due to its simple algorithmic structure and good convergence performance.
In network-consensus-based decentralized learning, a set of geographically distributed computing nodes collaborate to train a common learning model.
Each node holds a local dataset that may be too large to be sent to a centralized location due to network communication limits, or cannot be shared due to privacy/security risks.
A distinctive feature of network-consensus-baed decentralized learning is that there is a lack of a dedicated PS.
As a result, each node has to exchange information with its local neighbors to reach a consensus on a global optimal learning model.

Despite its growing significance in practice, the design of high-performance network-consensus-based decentralized learning faces three fundamental {\em conflicting} complexities, namely {\em sample, communication, and memory complexities}.
First, due to the high dimensionality of most deep learning models, it is impossible to leverage beyond first-order (stochastic) gradient information to compute the update direction in each iteration.
The variability of a stochastic gradient is strongly influenced by the number of training samples in its mini-batch.
However, the more training samples in a mini-batch, the higher computational cost of the stochastic gradient.
Second, by using fewer training samples in each iteration to trade for a lower computational cost, the resulting stochastic gradient unavoidably has a larger variance, which further leads to more iterations (hence communication rounds) to reach a certain training accuracy (i.e., slower convergence).
The low communication efficiency is particularly problematic in many wireless edge networks, where the communication links could be low-speed and highly unreliable.
Lastly, in many mobile edge-computing environments, the mobile devices could be severely limited by hardware resources (e.g., CPU/GPU, memory) and they cannot afford reserving a large memory space to run a very sophisticated decentralized learning algorithm that has too many intermediate variables.

Due to the above fundamental trade-off between sample, communication, and computing resource costs, the notions of sample, communication, and memory complexities (to be formally defined in Section~\ref{sec:related}) become three of the most important measures in assessing the performances of decentralized learning algorithms.
However, in the literature, most existing works have achieve low complexities in some of these measures, but not all (see Section~\ref{sec:related} for in-depth discussions).
The limitations of these existing works motivate to ask the following question: {\em Could we design a decentralized learning algorithm that strikes a good balance between sample complexity and communication complexity?}
In this paper, we answer the above question positively by proposing a new GT-STORM algorithm (\underline{g}radient-\underline{t}racking-based \underline{sto}chastic \underline{r}ecursive \underline{m}omentum) that achieves low sample, communication, and memory complexities.
Our main results and contributions are summarized as follows:

\begin{list}{\labelitemi}{\leftmargin=1em \itemindent=0.em \itemsep=.2em}

\item 
Unlike existing approaches, our proposed GT-STORM algorithm adopts a new estimator, which is updated with a consensus mixing of the neighboring estimators of the last iteration, which helps improve the global gradient estimation. 
Our method achieves the nice features of previous works \cite{tran2019hybrid,cutkosky2019momentum,di2016next,lu2019gnsd} while avoiding their pitfalls.
To some extent, our GT-STORM algorithm can be viewed as an indirect way of integrating the stochastic gradient method, variance reduction method, and gradient tracking method.

\item 
We provide a detailed convergence analysis and complexity analysis.
Under some mild assumptions and parameter conditions, our algorithm enjoys an $\tilde{O}(T^{-2/3})$ convergence rate. 
Note that this rate is much faster than the rate of $O(T^{-1/2})$ for the classic decentralized stochastic algorithms, e.g., DSGD \cite{jiang2017collaborative}, PSGD \cite{lian2017can} and GNSD \cite{lu2019gnsd}.
Also, we show that to reach an $\epsilon^2$-stationary solution, the total number of sample evaluations of our algorithm is $\tilde{O}(m^{1/2}\epsilon^{-3})$ and the communication round is $\tilde{O}(m^{-1/2}\epsilon^{-3})$.

\item 
We conduct extensive experiments to examine the performance of our algorithm, including both a non-convex logistic regression model on the LibSVM datasets and 
convolutional neural network models on MNIST and CIFAR-10 datasets. 
Our experiments show that the our algorithm outperforms two state-of-the-art decentralized learning algorithms \cite{lian2017can,lu2019gnsd}. 
These experiments corroborate our theoretical results.
 
\end{list}

The rest of the paper is organized as follows.
In Section~\ref{sec:related}, we first provide the preliminaries of network consensus optimization and discuss related works with a focus on sample, communication, and memory complexities.
In Section~\ref{Section: algorithm}, we present our proposed GT-STORM algorithm, as well as its communication, sample, and memory complexity analysis. 
We provide numerical results in Section~\ref{Section: experiment} to verify the theoretical results of our GT-STORM algorithm.
Lastly in Section \ref{Section: conclusion}, we provide concluding remarks.

\section{Preliminaries and Related Work} \label{sec:related}

To facilitate our technical discussions, in Section~\ref{sec:ncoa}, we first provide an overview on network consensus optimization and formally define the notions of sample, communication, and memory complexities of decentralized optimization algorithms for network consensus optimization.
Then, in Section~\ref{sec:sfoa}, we first review centralized stochastic first-order optimization algorithms for solving non-convex learning problems from a historical perspective and with a focus on sample, communication, and memory complexities.
Here, we introduce several acceleration techniques that motivate our GT-STORM algorithmic design.
Lastly, we review the recent developments of optimization algorithms for decentralized learning and compare them with our work.

\subsection{Network Consensus Optimization} \label{sec:ncoa}

As mentioned in Section~\ref{sec:Intro}, in decentralized learning, there are a set of geographically distributed computing nodes forming a network.
In this paper, we represent such a networked by an undirected connected network $\mathcal{G}=(\mathcal{N},\mathcal{L})$,
where $\mathcal{N}$ and $\mathcal{L}$ are the sets of nodes and edges, respectively, with $|\mathcal{N}| = m$. 
Each node can communicate with their neighbors via the edges in $\mathcal{L}$.
The goal of decentralized learning is to use the nodes to {\em distributively} and {\em collaboratively} solve a network-wide optimization problem as follows:
\begin{align}\label{Eq: general_problem}
\min_{\x \in \mathbb{R}^p} f(\x) = \min_{\x \in \mathbb{R}^p} \frac{1}{m}\sum_{i=1}^{m} f_i(\x),
\end{align}
where each local objective function $f_i(\x) \triangleq \Eb_{\zeta\sim\D_i} f_i(\x;\zeta)$ is only observable to node $i$ and not necessarily convex.
Here, $\D_i$ represents the distribution of the dataset at node $i$, and
$f_{i}(\x;\zeta)$ represents a loss function that evaluates the discrepancy between the learning model's output and the ground truth of a training sample $\zeta$.
To solve Problem~\eqref{Eq: general_problem} in a decentralized fashion, a common approach is to rewrite Problem (\ref{Eq: general_problem}) in the following equivalent form:
\begin{align}\label{Eq: consensus_problem}
& \text{Minimize} && \hspace{-.5in} \frac{1}{m}\sum_{i=1}^{m} f_i(\x_i) & \\
& \text{subject to} && \hspace{-.5in} \x_i = \x_j, && \hspace{-.5in} \forall (i,j) \in \mathcal{L}, \nonumber
\vspace{-.05in}
\end{align}  
where $\x \triangleq [\x_1^\top,\cdots,\x_m^\top]^\top$ and $\x_i$ is an introduced local copy at node $i$.
In Problem~\eqref{Eq: consensus_problem}, the constraints ensure that the local copies at all nodes are equal to each other, hence the term ``consensus.''
Thus, Problems~\eqref{Eq: general_problem} and \eqref{Eq: consensus_problem} share the same solutions.
The main goal of network consensus optimization is to design an algorithm to attain an $\epsilon^2$-stationary point $\x$ defined as follows:
\begin{align}\label{Eq: FOSP_network}
\underbrace{\Big\|\frac{1}{m}\sum_{i=1}^{m} \nabla f_i(\xb) \Big\|^2}_{\mathrm{Global \,\, gradient \,\, magnitude}} \!\!\!\! + \underbrace{\frac{1}{m}\sum_{i=1}^{m}\|\x_{i}- \xb\|^2}_{\mathrm{Consensus \,\, error}} \le \epsilon^2,
\end{align}
where $\xb \triangleq \frac{1}{m}\sum_{i=1}^{m} \x_{i}$ denotes the global average across all nodes.
Different from the traditional $\epsilon^2$-stationary point in centralized optimization problems, the metric in Eq.~\eqref{Eq: FOSP_network} has two terms: the first term is the gradient magnitude for the (non-convex) global objective and the second term is the average consensus error of all local copies.
To date, many decentralized algorithms have been developed to compute the $\epsilon^2$-stationary point (see Section~\ref{sec:sfoa}).
However, most of these algorithms suffer limitations in sample, communication, and memory complexities.
In what follows, we formally state the definitions of sample, communication, and memory complexities used in the literature (see, e.g., \cite{sun2019improving}):

\begin{defn}[Sample Complexity]
The sample complexity is defined as the total number of the incremental first-order oracle (IFO) calls required across all the nodes to find an $\epsilon^2$-stationary point defined in Eq.~(\ref{Eq: FOSP_network}), where one IFO call evaluates a pair of $(f_i(\x;\zeta), \nabla f_i(\x;\zeta))$ on a sample $\zeta \sim \D_i$ and parameter $\x \in \mathbb{R}^p$ at node $i.$
\end{defn}

\begin{defn}[Communication Complexity]
The communication complexity is defined as the total rounds of communications required to find an $\epsilon^2$-stationary point defined in Eq.~(\ref{Eq: FOSP_network}), where each node can send and receive a $p$-dimensional vector with its neighboring nodes in one communication round.
\end{defn}

\begin{defn}[Memory Complexity]
The memory complexity is defined as total dimensionality of all intermediate variables in the algorithm run by a node to find an $\epsilon^2$-stationary point in Eq.~(\ref{Eq: FOSP_network}).
\end{defn}

To make sense of these three complexity metrics into perspective, consider the standard centralized gradient descent (GD) method as an example.
Note that the GD algorithm has an $O(1/T)$ convergence rate for non-convex optimization, which suggests $O(\epsilon^{-2})$ communication complexity.
Also, it takes a full gradient evaluation in each iteration, i.e., $O(n)$ per-iteration sample complexity, where $n$ is the total number of samples.
This implies $O(n\epsilon^{-2})$ sample complexity to converge to an $\epsilon^{2}$-stationary point.
Hence, the sample complexity of GD is high if the dataset size $n$ is large.

In contrast, consider the classical stochastic gradient descent (SGD) algorithm that is widely used in machine learning.
The basic idea of SGD is to lower the gradient evaluation cost by using only a mini-batch of samples in each iteration.
However, due to the sample randomness in mini-batches, the convergence rate of SGD for non-convex optimization is reduced to $O(1/\sqrt{T})$~\cite{ghadimi2013stochastic,bottou2018optimization,zhou2018new}.
Thus, to reach an $\epsilon^2$-stationary point $\x$ with $\|\nabla f(\x)\|^2 \le \epsilon^2$, SGD has $O(\epsilon^{-4})$ sample complexity, which could be either higher or lower than the $O(n\epsilon^{-2})$ sample complexity of the GD method, depending on the relationship between $n$ and $\epsilon$.
Also, for $p$-dimensional problems, both GD and SGD have memory complexity $p$, since they only need a $p$-dimensional vector to store (stochastic) gradients. 

\subsection{Related Work} \label{sec:sfoa}

{\bf 1) Centralized First-Order Methods with Low Complexities:}
Now, we review several state-of-the-art low-complexity centralized stochastic first-order methods that are related to our GT-STORM algorithm.
To reduce the overall sample and communication complexities of the standard GD and SGD algorithms, a natural approach is variance reduction.
Earlier works following this approach include SVRG \cite{johnson2013accelerating,reddi2016stochastic}, SAGA \cite{defazio2014saga} and SCSG \cite{lei2017non}. 
These algorithms has an overall sample complexity of $O(n + n^{2/3}\epsilon^{-2})$.
A more recent variance reduction method is the stochastic path-integrated differential estimator (SPIDER)~\cite{fang2018spider}, which is based on the SARAH gradient estimator developed by Nguyen {\em et al.} \cite{nguyen2017sarah}.
SPIDER further lowers the sample complexity to $O(n + \sqrt{n}\epsilon^{-2})$, which attains the $\Omega(\sqrt{n}\epsilon^{-2})$ theoretical lower bound for finding an $\epsilon^2$-stationary point for $n = O(\epsilon^{-4})$.
More recently, to improve the small step-size $O(\epsilon L^{-1})$ in SPIDER, a variant called SpiderBoost was proposed in~\cite{wang2019spiderboost}, which allows a larger constant step-size $O(L^{-1})$ while keeping the same $O(n + \sqrt{n}\epsilon^{-2})$ sample complexity.
It should be noted, however, that the significantly improved sample complexity of SPIDER/SpiderBoost is due to a restrictive assumption that a universal Lipschitz smoothness constant exists for all local objectives $f(\cdot;\zeta_i)$ $\forall i$.
This means that the objectives are ``similar'' and there are no ``outliers'' in the training samples.
Meanwhile, to obtain the optimal communication complexity, SpiderBoost require a (nearly) full gradient every $\sqrt{n}$ iterations and a mini-batch of stochastic gradient evaluation with batch size $\sqrt{n}$ in each iteration. 

To overcome the above limitations, a hybrid stochastic gradient descent (Hybrid-SGD) method is recently proposed in~\cite{tran2019hybrid}, where a convex combination of the SARAH estimator~\cite{nguyen2017sarah} and an unbiased stochastic gradient is used as the gradient estimator.
The Hybrid-SGD method relaxes the universal Lipschitz constant assumption in SpiderBoost to an average Lipschitz smoothness assumption.
Moreover, it only requires two samples to evaluate the gradient per iteration.
As a result, Hybrid-SGD has a $O(\epsilon^{-3})$ sample complexity that is {\em independent} of dataset size. 
Although Hybrid-SGD is for centralized optimization, the interesting ideas therein motivate our GT-STORM approach for {\em decentralized learning} following a similar token.
Interestingly, we show that in decentralized settings, our GT-STORM method can further improve the gradient evaluation to only {\em one sample} per iteration, while not degrading the communication complexity order.
Lastly, we remark that all algorithms above have memory complexity at least $2p$ for $p$-dimensional problems.
In contrast, GT-STORM enjoys a $p$ memory complexity.

\vspace{.05in}
{\bf 2) Decentralized Optimization Algorithms} 
In the literature, many decentralized learning optimization algorithms have been proposed to solve Problem~(\ref{Eq: general_problem}), e.g., first-order methods \cite{nedic2009distributed,yuan2016convergence,shi2015extra, di2016next}, prime-dual methods \cite{sun2019distributed,mota2013d}, Newton-type methods \cite{mokhtari2016decentralized,eisen2017decentralized} (see in~\cite{nedic2018network,chang2020distributed} for comprehensive surveys).
In this paper, we consider decentralized first-order methods for the non-convex network consensus optimization in~\eqref{Eq: consensus_problem}.
In the literature, the convergence rate of the well-known decentralized gradient descent (DGD) algorithm \cite{nedic2009distributed} was studied in~\cite{zeng2018nonconvex}, which showed that DGD with a constant step-size converges with an $O(1/T)$ rate to a step-size-dependent error ball around a stationary point.
Later, a gradient tracking (GT) method was proposed in~\cite{di2016next} to find an $\epsilon^2$-stationary point with an $O(1/T)$ convergence rate under constant step-sizes. 
However, these methods require a full gradient evaluation per iteration, which yields $O(n\epsilon^{-2})$ sample complexity.
To reduce the per-iteration sample complexity, stochastic gradients are adopted in the decentralized optimization, e.g., DSGD \cite{jiang2017collaborative}, PSGD \cite{lian2017can}, GNSD \cite{lu2019gnsd}.
Due to the randomness in stochastic gradients, the convergence rate is reduced to $O(1/\sqrt{T}).$
Thus, the sample and communication complexities of these stochastic methods are $O(\epsilon^{-4})$ and $O(m^{-1}\epsilon^{-4})$, two orders of magnitude higher than their deterministic counterparts.
To overcome the limitations in stochastic methods, a natural idea is to use  variance reduction techniques similar to those for centralized optimization to reduce the sample and communication complexities for the non-convex network consensus optimization.
So far, existing works on the decentralized stochastic variance reduction methods include DSA~\cite{mokhtari2016dsa}, diffusion-AVRG~\cite{yuan2018variance} and GT-SAGA~\cite{xin2019variance} etc., all of which focus on convex problems.
To our knowledge, the decentralized gradient estimation and tracking (D-GET) algorithm in~\cite{sun2019improving} is the only work for non-convex optimization.
D-GET integrates the decentralized gradient tracking~\cite{lu2019gnsd} and the SpiderBoost gradient estimator~\cite{wang2019spiderboost} to obtain $O(mn+m\sqrt{n}\epsilon^{-2})$ {\em dataset-size-dependent} sample complexity and $O(\epsilon^{-2})$ communication complexity.
Recall that the sample and communication complexities of GT-STORM are $O(m^{1/2}\epsilon^{-3})$ and $O(m^{-1/2}\epsilon^{-3})$, respectively.
Thus, if dataset size $n=\Omega(\epsilon^{-2})$, D-GET has a higher sample complexity than GT-STORM.
As an example, when $\epsilon=10^{-2}$, $n$ is on the order of $10^4$, which is  common in modern machine learning datasets.
Also, the memory complexity of D-GET is $2p$ as opposed to the $p$ memory complexity of GT-STORM.
This implies a huge saving with GT-STORM if $p$ is large, e.g., $p\approx 10^6$ in many deep learning models.

\section{A Gradient-Tracking Stochastic Recursive Momentum Algorithm}\label{Section: algorithm}

In this section, we introduce our \underline{g}radient-\underline{t}racking-based \underline{sto}chastic \underline{r}ecursive \underline{m}omentum (GT-STORM) algorithm for solving Problem (\ref{Eq: consensus_problem}) in Section~\ref{sec:gt-storm}.
Then, we will state the main theoretical results and their proofs in Sections~\ref{sec:main_results} and~\ref{sec:proofs}, respectively.

\subsection{The GT-STORM Algorithm} \label{sec:gt-storm}

In the literature, a standard starting point to solve Problem~\eqref{Eq: consensus_problem} is to reformulate the problem as \cite{nedic2009distributed}:
\vspace{-.03in}
\begin{align} \label{Eq:DGD_reformulation}
& \text{Minimize} && \hspace{-.5in} \frac{1}{m}\sum_{i=1}^{m} f_i(\x_i) & \\
& \text{subject to} && \hspace{-.5in} (\W \otimes \I_{p}) \x = \x, &&\nonumber
\vspace{-.09in}
\end{align}
where $\I_{p}$ denotes the $p$-dimensional identity matrix, the operator $\otimes$ denotes the Kronecker product, and $\W\in \mathbb{R}^{m\times m}$ is often referred to as the consensus matrix.
We let $[\W]_{ij}$ represent the element in the $i$-th row and the $j$-th column in $\W$.
For Problems~\eqref{Eq:DGD_reformulation} and~\eqref{Eq: consensus_problem} to be equivalent, $\W$ should satisfy the following properties:
\begin{enumerate}[topsep=1pt, itemsep=-.1ex, leftmargin=.25in]
\item[(a)] {\em Doubly Stochastic:} $\sum_{i=1}^{m} [\mathbf{W}]_{ij}=\sum_{j=1}^{m} [\mathbf{W}]_{ij}=1$.
\item[(b)] {\em Symmetric:} $[\mathbf{W}]_{ij} = [\W]_{ji}$, $\forall i,j \in \mathcal{N}$. 
\item[(c)] {\em Network-Defined Sparsity Pattern:} $[\W]_{ij} > 0$ if $(i,j)\in \mathcal{L};$ otherwise $[\mathbf{W}]_{ij}=0$, $\forall i,j \in \mathcal{N}$.
\end{enumerate}
The above properties imply that the eigenvalues of $\W$ are real and can be sorted as $-1 < \lambda_m \leq \cdots \leq \lambda_2 < \lambda_1 = 1$.
We define the second-largest eigenvalue in magnitude of $\W$ as $\lambda \triangleq \max\{|\lambda_2|,|\lambda_m|\}$ for the further notation convenience.
It can be seen later that $\lambda$ plays an important role in the step-size selection and the algorithm's convergence rate.

As mentioned in Section~\ref{sec:ncoa}, our GT-STORM algorithm is inspired by the GT method \cite{di2016next,nedich2016geometrically} for reducing consensus error and the recursive variance reduction (VR) methods \cite{fang2018spider,wang2019spiderboost} developed for centralized optimization.
Specifically, in the centralized GT method, an estimator $\y$ is introduced to track the global gradient:
\begin{align}\label{Eq: GT}
\y_{t} = \W\y_{t-1} + \g_{t} - \g_{t-1} ,
\end{align}
where $\g_t$ is the gradient estimation in the $t$th iteration.
Meanwhile, to reduce the stochastic error, a gradient estimator $\v$ in VR methods is updated recursively based on a {\em double-loop} structure as follows:
\begin{align}\label{Eq: VR}
\v_{t} = \v_{t-1} + \nabla f(\x_t; \zeta_t) - \nabla f(\x_{t-1}; \zeta_t), \quad \text{if } \text{mod}(t, q) \neq 0,
\end{align}
where $\nabla f(\x; \zeta)$ is the stochastic gradient dependent on parameter $\x$ and a data sample $\zeta,$ and $q$ is the number of the inner loop iterations.
On the other hand, if $\text{mod}(t,q) = 0$, $\v_{t}$ takes a full gradient.
Note that these two estimators have a {\em similar} structure: Both are {\em recursively} updating the previous estimation based on the difference of the gradient estimations between two consecutive iterations (i.e., momentum).
This motivates us to consider the following question:
{\em{ Could we somehow ``integrate'' these two methods to develop a new decentralized gradient estimator to track the global gradient and reduce the stochastic error at the same time?}}
Unfortunately, the GT and VR estimators can not be combined straightforwardly.
The major challenge lies in the structural difference in the outer loop iteration (i.e., $\text{mod}(t,q) = 0$), where the VR estimator requires full gradient and does not follow the recursive updating structure.

Surprisingly, in this paper, we show that there exists an ``indirect'' way to achieve the salient features of both GT and VR.
Our approach is to abandon the double-loop structure of VR and pursue a {\em single-loop} structure.
Yet, this single-loop structure should still be able to reduce the variance and consistently track the global gradient.
Specifically, we introduce a parameter $\beta_t \in [0,1]$ in the recursive update and integrate it with a consensus step as follows:
\begin{align}
\v_{i,t} \!=\! \beta_t & \sum\nolimits_{j \in \mathcal{N}_{i}} [\W]_{ij} \v_{j,t-1}  \!+\! \nabla f_i(\x_{i,t};\zeta_{i,t}) \!-\! \beta_t \nabla f_i(\x_{i,t-1};\zeta_{i,t}), \!\!\!
\end{align}
where $\x_{i,t},$ $\v_{i,t}$ and $\zeta_{i,t}$ are the parameter, gradient estimator, and random sample in the $t$th iteration at node $i$, respectively.
Note that the estimator reduces to the classical stochastic gradient estimator when $\beta_t = 0$.
On the other hand, if we set $\beta_t = 1$, the estimator becomes the (stochastic) gradient tracking estimator based on a single sample (implying low sample complexity).
Then, the key to the success of our GT-STORM design lies in meticulously choosing parameter $\beta_t$ to mimic the gradient estimator technique in centralized optimization~\cite{cutkosky2019momentum,tran2019hybrid}.
Lastly, the local parameters can be updated by the conventional decentralized stochastic gradient descent step:
\begin{align} \label{eqn:main_update}
\x_{i,t+1} = \sum\nolimits_{j \in \mathcal{N}_{i}} [\W]_{ij} \x_{j,t} - \eta_t \v_{i,t},
\end{align}
where $\eta_t$ is the step-size in iteration $t$. 
To summarize, we state our algorithm in Algorithm~1 as follows.


\medskip
\hrule 
\vspace{.03in}
\noindent {\bf Algorithm~1:} Gradient-Tracking-based Stochastic Recursive Momentum Algorithm (GT-STORM).
\vspace{.03in}
\hrule
\vspace{0.1in}
\noindent {\bf Initialization:}
\begin{enumerate} [topsep=1pt, itemsep=-.1ex, leftmargin=.2in]
\item[1.] Choose $T>0$ and let $t=1$. Set $\x_{i,0} = \x^0$ at node $i$. Calculate $\v_{i,0} = \nabla f_i(\x_{i,0};\zeta_{i,0})$ at node $i$. 
\end{enumerate}

\noindent {\bf Main Loop:}
\begin{enumerate} [topsep=1pt, itemsep=-.1ex, leftmargin=.2in]
\item[2.] In the $t$-th iteration, each node sends $\x_{i,t-1}$ and local gradient estimator $\v_{i,t-1}$ to its neighbors. Meanwhile, upon the reception of all neighbors' information, each node performs the following:
    \begin{enumerate} [topsep=1pt, itemsep=-.1ex, leftmargin=.18in]
	\item[a)] Update local parameter: $\x_{i,t} = \sum\nolimits_{j \in \mathcal{N}_{i}} [\W]_{ij} \x_{j,t-1} - \eta_{t-1} \v_{i,t-1}$.
	\item[b)] Update local gradient estimator: $\v_{i,t} = \beta_{t} \sum\nolimits_{j \in \mathcal{N}_{i}} [\W]_{ij} \v_{j,t-1}$ $+ \nabla f_{i}(\x_{i,t};\zeta_{i,t}) -\beta_{t} \nabla f_{i}(\x_{i,t-1};\zeta_{i,t})$.
\end{enumerate}

\item[3.] Stop if $t>T$; otherwise, let $t \leftarrow t+1$ and go to Step 2. 
\end{enumerate}
\smallskip
\hrule
\medskip

Two remarks for Algorithm~1 are in order.
First, thanks to the single-loop structure, GT-STORM is easier to implement compared to the low-sample-complexity D-GET~\cite{sun2019improving} method, which has in a double-loop structure.
Second, GT-STORM only requires $p$ memory space due to the use of only one intermediate vector $\v$ at each node.
In contrast, the memory complexity of D-GET is $2p$ (cf. $\y$ and $\v$ in \cite{sun2019improving}).
This 50\% saving is huge particularly for deep learning models, where the number of parameters could be in the range of millions.

\subsection{Main Theoretical Results}\label{sec:main_results}

In this section, we will establish the complexity properties of the proposed GT-STORM algorithm.
For better readability, we state the main theorem and its corollary in this section and provide the intermediate lemmas to Section~\ref{sec:proofs}.
We start with the following assumptions on the global and local objectives:
\begin{assum}\label{Assump: obj}
The objective function $f(\x) = \frac{1}{m}\sum_{i=1}^{m} f_i(\x)$ with $f_i(\x) = \Eb_{\zeta\sim\D_i} f_i(\x;\zeta)$ satisfies the following assumptions:
\begin{enumerate}[topsep=1pt, itemsep=-.1ex, leftmargin=.25in]
	\item[(a)] {\em Boundedness from below:} There exists a finite lower bound $f^* = \inf_\x f(x) > -\infty;$
	\item[(b)] {\em $L$-average smoothness:} $f_i(\cdot;\zeta_i)$ is $L$-average smooth on $\mathbb{R}^p$, i.e., there exists a positive constant $L,$ such that $\Eb_{\zeta\!\sim\!\D_i}[\|\nabla f_i(\x;\zeta)\!-\! \nabla f_i(\y;\zeta)\|^2]\! \le \!L^2\|\x\!-\!\y\|^2, \forall \x,\y \in \mathbb{R}^p, i \in [m]$;
	\item[(c)] {\em Bounded variance:} There exists a constant $\sigma\ge 0$ such that $\Eb_{\zeta \sim \D_i}[\|\nabla f_i(\x;\zeta) - \nabla f_i(\x)\|^2] \le \sigma^2, \forall \x \in \mathbb{R}^p, i\in[m]$;
	\item[(d)] {\em Bounded gradient:} There exists a constant $G\ge 0$ such that $\Eb_{\zeta \sim\D_i}[\|\nabla f_i(\x;\zeta)\|^2] \le G^2, \forall \x \in \mathbb{R}^p, i \in [m]$.
\end{enumerate}
\end{assum}
In the above assumptions, (a) and (c) are standard in the stochastic non-convex optimization literature;
(b) is an expected Lipschitz smoothness condition over the data distribution, which implies the conventional global Lipschitz smoothness \cite{ghadimi2013stochastic} by the Jensen's inequality. 
Note that (b) is weaker than the individual Lipschitz smoothness in \cite{fang2018spider,wang2019spiderboost,sun2019improving}: if there exists an outlier data sample, then the individual objective function might have a very large smoothness parameter while the average smoothness can still be small;
(d) is equivalent to the Lipschitz continuity assumption, which is also commonly used for non-convex stochastic algorithms \cite{zhou2018generalization,karimireddy2019error,koloskova2019decentralized} and is essential for analyzing the decentralized gradient descent method \cite{yuan2016convergence,zeng2018nonconvex,jiang2017collaborative}.\footnote{Note that under the assumption (b), as long as the parameter $\x$ is bounded, (d) is satisfied.}

For convenience, in the subsequent analysis, we define $\tilde{\W} = \W \otimes \I_m,$ $\g_{i,t} = \nabla f_i(\x_{i,t}),$ $\u_{i,t} = \nabla f_i(\x_{i,t};\zeta_{i,t})$, $\w_{i,t} = \nabla f_i(\x_{i,t};\zeta_{i,t}) - \nabla f_i(\x_{i,t-1};\zeta_{i,t})$ and $\a_{t} = [\a_{1,t}^\top,\cdots,\a_{m,t}^\top]^\top$ and $\bar{\a}_t =\frac{1}{m} \sum_{i=1}^{m}\a_{i,t},$ for $\a \in \{\x,\u,\w,\v,\g\}$.  
Then, the algorithm can be compactly rewritten in the following matrix-vector form:
\begin{align}
\x_t &= \tilde{\W}\x_{t-1} -\eta_{t-1} \v_{t-1},\label{Eq: alg_updating_x}\\
\v_t &= \beta_t \tilde{\W}\v_{t-1} + \beta_t\w_{t} + (1-\beta_t)\u_{t}\label{Eq: alg_updating_v}.
\end{align}
Furthermore, since $\1^\top\W = \1^\top,$ we have $\xb_t = \xb_{t-1} - \eta_{t-1} \bv_{t-1},$ $\bv_{t} = \beta_t {\bv}_{t-1} +\beta_t\bar{\w}_t+(1-\beta_t)\bar{\u}_{t}.$
We first state the convergence result for Algorithm~1 as follows:
\begin{thm}\label{Theorem: convergence1}
Under Assumption~1 and with the positive constants $c_0$ and $c_1$ satisfying 
$1-( 1+c_1)\lambda^2 - \frac{1}{c_0} > 0$,
if we set $\eta_t = \tau/(\omega + t)^{1/3}$ and $\beta_{t+1} = 1- \rho\eta_t^2$, with $\tau >0,$ $\omega \ge \max\{2,\tau^3/\min\{k_1^3,k_2^3,k_3^3\}\}$ and $\rho = 2/(3\tau^3) + 32 L^2$,
then we have the following result for Algorithm~1:
\begin{align}\label{Eq: convergence1}
&\min_{t\in[T]} \Eb[\|\nabla f(\xb_{t})\|^2]+ \frac{1}{m}\Eb[\|\x_{t}-\1\otimes\xb_{t}\|^2] \notag\\
\le
&
\frac{2(f(\xb_{0})  - f(\xb^*))}{\tau{(T+1)}^{2/3}}  
+ \frac{2c_0\Eb[\|\v_{0}-\1\otimes\bv_{0}\|^2]}{m\tau{(T+1)}^{2/3}} \notag\\
&
+ \frac{(\omega - 1)\sigma^2}{16mL^2\tau^2{(T+1)}^{2/3}} 
+ \frac{\rho^2\sigma^2\ln(\omega+T-1)}{8mL^2(T+1)^{2/3}}\notag\\
&
+ \frac{12(1 + \frac{1}{c_1})c_0\tau^{1/3}G^2\rho^2}{(\omega - 1)^{1/3}(T+1)^{2/3}} + O\Big(\frac{c_3\omega}{\tau T^{5/3}}\Big),
\end{align}
where $c_3 = \max\{1,\omega/(m\tau^2), \tau^{4/3}/\omega^{1/3}, \tau \ln(\omega +T)/m\},$ and the constants $k_1,$ $k_2$ and $k_3$ are:
\begin{align}
k_1 &= 1/\Big(2L + 32(1 + \frac{1}{c_1})c_0L^2\Big),\\
k_2 &= \Big(1- (1+c_1)\lambda^2\Big) / \Big(1+\frac{1}{c_1} + \frac{1}{c_0}\Big) ,\\
k_3 &=\sqrt{\Big(1-( 1+c_1)\lambda^2 - \frac{1}{c_0}\Big)/\Big(\frac{2}{3\tau^3} + \frac{2L^2+1}{2c_0}\Big)}.
\end{align}
\end{thm}
In Theorem \ref{Theorem: convergence1}, $c_0$ and $c_1$ are two constants depending on the network topology, which in turn will affect the step-size and convergence: with a sparse network, i.e., $\lambda$ is close to but not exactly one (recall that $\lambda = \max\{|\lambda_2|,|\lambda_m|\}$). 
In order for $1-( 1+c_1)\lambda^2 - \frac{1}{c_0} > 0$ to hold, $c_0$ needs to be large and $c_1$ needs be close to zero, which leads to small $k_1,$ $k_2$ and $k_3.$
Note that the step-size $\eta_t$ is of the order $O(t^{-1/3}),$ which is larger than the $O(t^{-1/2})$ order for the classical decentralized SGD algorithms.
With this larger step-size, the convergence rate is $O(t^{-2/3})$ and faster than the rate $O(t^{-1/2})$ for the decentralized SGD algorithms.
Based on Theorem~\ref{Theorem: convergence1}, we have the sample and communication complexity results for Algorithm~1:
\begin{cor}\label{Cor: complexity}
Under the conditions in Theorem~\ref{Theorem: convergence1},
if $\tau = O(m^{1/3})$ and $\omega = O(m^{4/3})$, 
then to achieve an $\epsilon^2$-stationary solution,
the total communication rounds are on the order of $\tilde{O}(m^{-1/2}\epsilon^{-3})$ and the total samples evaluated across the network is on the order of $\tilde{O}(m^{1/2}\epsilon^{-3}).$
\end{cor}


\subsection{Proofs of the Theoretical Results} \label{sec:proofs}
 
Due to space limitation, we provide a proof sketch for Theorem~\ref{Theorem: convergence1} here and relegate the details to the appendices.
First, we bound the error of gradient estimator $\Eb[\|\v_t - \g_t \|^2 ]$ as follows: 

\begin{lem}[Error of Gradient Estimator]\label{Lemma: Error of v}
Under Assumption \ref{Assump: obj} and with $\v_t$ defined in (\ref{Eq: alg_updating_v}), it holds that $\Eb[\|\bv_t - \bg_t \|^2 ] \le \beta_t^2 \Eb[\|\bv_{t-1} - \bg_{t-1}\|^2] + \frac{2\beta_t^2L^2}{m}\Eb[\|\x_{t}-\x_{t-1}\|^2] 
+ \frac{2(1-\beta_t)^2\sigma^2}{m}$.
\end{lem}
It can be seen that the upper bound depends on the error in the previous step with a factor $\beta_t^2$. 
This will be helpful when we construct a potential function.
Then, according to the algorithm updates (\ref{Eq: alg_updating_x})--(\ref{Eq: alg_updating_v}), we show the following descent inequality: 

\begin{lem}[Descent Lemma]\label{Lemma: Descend lemma}
Under Assumption \ref{Assump: obj}, Algorithm~1 satisfies:
$\Eb[f(\xb_{t+1})]-\Eb[f(\xb_{t})] \le  - \frac{\eta_t}{2}\Eb[\|\nabla f(\xb_{t})\|^2] - (\frac{\eta_t}{2}  - \frac{L\eta_t^2}{2}) \times$ $\Eb[\|\bv_t\|^2] + \eta_t\Eb[\|\bv_t - \bg_t\|^2] + \frac{L^2\eta_t}{m}\Eb[\|\x_t-\1\otimes\xb_{t} \|^2 ]$.
\end{lem}
We remark that the right-hand-side (RHS) of the above inequality contains the consensus error of local parameters $\sum_{t= 0}^{T}\Eb[\|\x_t-\1\otimes\xb_{t} \|^2 ]$, which makes the analysis more difficult than that of the centralized optimization.
Next, we prove the contraction of iterations in the following lemma, which is useful in analyzing the decentralized gradient tracking algorithms.

\begin{lem}[Iterates Contraction]\label{Lemma: Iterates Contraction}
The following contraction properties of the iterates produced by Algorithm~1 hold:
\begin{align}
\|\x_{t+1}-\1\otimes\xb_{t+1}\|^2  \le (1+c_1)&\lambda^2\|\x_{t} -\1\otimes\xb_{t} \|^2 \notag\\
&+ (1+\frac{1}{c_1}) \eta_{t}^2\|\v_{t}-\1 \otimes \bv_{t}\|^2, 
\end{align}
\vspace{-.2in}
\begin{align}
\|\v_{t+1}-\1&\otimes\bv_{t+1}\|^2  \le
(1+c_1)\beta_{t+1}^2\lambda^2\|\v_{t} - \1 \otimes\bv_{t}\|^2 \notag\\
& + 2(1 + \frac{1}{c_1})\big(\beta_{t+1}^2\|\w_{t+1}\|^2 + (1-\beta_{t+1})^2\|\u_{t+1}\|^2\big), 
\end{align}
where $c_1$ is a positive constant.
Additionally, we have 
\begin{align}
\|\x_{t+1}-\x_{t}\|^2 
 \le 
8\|(\x_{t} &- \1\otimes\xb_{t}) \|^2 \notag\\
&+ 4\eta_t^2 \|\v_{t} - \1\otimes\bv_{t}\|^2 + 4\eta_t^2m\|\bv_{t}\|^2.
\end{align}
\end{lem}

Finally, we define a potential function in (\ref{Eq: potential function}), based on which we prove the convergence bound:

\begin{lem}(Convergence of Potential Function)\label{Lemma: Potential Func}
Define the following potential function:
\begin{align}\label{Eq: potential function}
H_t = \Eb[f(\xb_{t})+ \frac{1}{32L^2\eta_{t-1}}\|\bg_{t} - \bv_{t}\|^2& + \frac{c_0}{m\eta_{t-1}}\|\x_{t}-\1\otimes\xb_{t}\|^2 \notag\\
&+ \frac{c_0}{m}\|\v_{t}-\1\otimes\bv_{t}\|^2],
\end{align}
where $c_0$ is a positive constant.
Under Assumption \ref{Assump: obj}, if we set $\eta_t = \tau/(\omega + t)^{1/3}$ and $\beta_{t+1} = 1- \rho\eta_t^2$, where $\tau,$ $\omega \ge 2,$ $\rho = 2/(3\tau^3) + 32 L^2$ are three constants,
then it holds that:
\begin{align}
H_{t+1} -H_t 
\le
&
 - \frac{\eta_t}{2}\Eb[\|\nabla f(\xb_{t})\|^2] + \frac{\rho^2\sigma^2\eta_t^3}{16mL^2}
+ 2(1 + \frac{1}{c_1})c_0G^2\rho^2\eta_t^4 \notag\\
&
- \frac{c_0C_1}{m\eta_t}[\|\x_{t}-\1\otimes\xb_{t}\|^2] 
- \frac{c_0C_2}{m}\Eb[\|\v_{t} - \1 \otimes\bv_{t}\|^2] \notag\\
&
 - \frac{C_3\eta_t}{4}\Eb[\|\bv_t\|^2],
\end{align}
where $C_1,$ $C_2,$ and $C_3$ are following constants:
$C_1 = 1-( 1+c_1)\lambda^2 - \frac{1}{2c_0} - 16(1 + \frac{1}{c_1})L^2\eta_t -  \Big(\frac{2}{3\tau^3} + \frac{L^2}{c_0}\Big)\eta_t^2,$ 
$C_2  = 1- (1+c_1)\lambda^2 - (1+\frac{1}{c_1})\eta_t - \frac{\eta_t}{4c_0} - 8(1 + \frac{1}{c_1})L^2\eta_{t}^2$, 
$C_3  = 1  - 2L\eta_t - 32(1 + \frac{1}{c_1})c_0L^2\eta_t$.
\end{lem}

Finally, by properly selecting the parameters, constants $C_1,$ $C_2$ and $C_3$ can be made non-negative, which leads to Theorem \ref{Theorem: convergence1}. 

\section{Experimental Results}\label{Section: experiment}

In this section, we conduct experiments using several non-convex machine learning problems to evaluate the performance of our method.
In particular, we compare our algorithm with the following state-of-art {\em single-loop} algorithms:
\begin{list}{\labelitemi}{\leftmargin=1em \itemindent=0.em \itemsep=.2em}
	\item DSGD \cite{nedic2009distributed,yuan2016convergence,jiang2017collaborative}: Each node performs: $\x_{i,t+1} = \sum_{j \in \mathcal{N}_{i}}$ $[\W]_{ij}\x_{j,t} - \eta \nabla f_i (\x_{i,t}; \zeta_{i,t})$, where the stochastic gradient $\nabla f_i (\x_{i,t}; \zeta_{i,t})$ corresponds to random sample $\zeta_{i,t}$. Then, each node exchanges the local parameter $\x_{i,t}$ with its neighbors. 
	\item GNSD \cite{lu2019gnsd}: Each node keeps two variables $\x_{i,t}$ and $\y_{i,t}$. The local parameter $\x_{i,t}$ is updated as $\x_{i,t+1} \!=\! \sum_{j \in \mathcal{N}_{i}} [\W]_{ij}\x_{j,t}\! -\! \eta \y_{i,t}$ and the tracked gradient $\y_{i,t}$ is updated as $\y_{i,t+1} \!=\! \sum_{j \in \mathcal{N}_{i}} [\W]_{ij}\y_{j,t} \! + \! \nabla f_i (\x_{i,t+1}; \zeta_{i,t+1})\! - \!\nabla f_i (\x_{i,t}; \zeta_{i,t}).$
\end{list}
Here, we compare with the above two classes of stochastic algorithms because they all employ a single-loop structure and do not require full gradient evaluations.
We note that it is hard to have a direct and fair comparison with D-GET~\cite{sun2019improving} numerically, since D-GET uses full gradients and has a double-loop structure.

\smallskip
{\em \underline{Network Model:}} 
The communication graph $\mathcal{G}$ is generated by the Erd$\ddot{\text{o}}$s-R$\grave{\text{e}}$nyi graph with different edge connectivity probability $p_c$ and number of nodes $m$.
We set $m=10$ and the edge connectivity probability as $p_c =0.5$. 
The consensus matrix is chosen as $\W = \I - \frac{2}{3\lambda_{\text{max}}(\mathbf{L})} \mathbf{L},$ where $\mathbf{L}$ is the Laplacian matrix of $\mathcal{G}$, and $\lambda_{\text{max}}(\mathbf{L})$ denotes the largest eigenvalue of $\mathbf{L}$.

\smallskip
{\bf 1) Non-convex logistic regression:} 
In our first experiment, we consider the binary logistic regression problem with a non-convex regularizer \cite{wang2018cubic,wang2019spiderboost,tran2019hybrid}:
\begin{align}
\min_{\x \in \mathbb{R}^d} -\frac{1}{mn} &\sum_{i=1}^{m}\sum_{j=1}^{n} [y_{ij}\log \big(\frac{1}{1+e^{-\x^\top\zeta_{ij}}} \big) + \notag\\
&(1-y_{ij}) \log \big(\frac{e^{-\x^\top\zeta_{ij}}}{1+e^{-\x^\top\zeta_{ij}}}\big)]+ \alpha \sum_{i=1}^{d} \frac{\x_i^2}{1+\x_i^2},
\end{align}
where the label $y_{ij} \in \{0,1\},$ the feature $\zeta_{ij} \in \mathbb{R}^{d}$ and $\alpha =0.1$.


\smallskip
{\em 1-a) \underline{Datasets:}} We consider three commonly used binary classification datasets from LibSVM: $a9a$, $rcv1.binary$ and $ijcnn1$. The $a9a$ dataset has $32561$ samples, $123$ features, the $rcv1.binary$ dataset has $20242$ samples, $47236$ features, and the $ijcnn1$ dataset has $49990$ samples, $22$ features. We evenly divide the dataset into $m$ sub-datasets corresponding to the $m$ nodes.

\smallskip
{\em 1-b) \underline{Parameters:}}  
For all algorithms, we set the batch size as one and the initial step-size $\eta_0$ is tuned by searching over the grid $\{0.01, 0.02, 0.05 , 0.1, 0.2,  0.5, 1.0\}.$ 
For DSGD and GNSD, the step-size is set to $\eta_t = \eta_0/\sqrt{1+0.1 t}$, which is on the order of $O(t^{-1/2})$ following the state-of-the-art theoretical result \cite{lu2019gnsd}. 
For GT-STORM, the step-size is set as $\eta_t = \eta_0/\sqrt[3]{1+0.1 t}$, which is on the order of $O(t^{-1/3})$ as specified in our theoretical result.
In addition, we choose the parameter $\rho $ for GT-STORM as $1/\eta_0^2$, so that $\beta_1 = 0$ in the first step. 

\begin{figure*}[t!]
 \begin{minipage}[t]{0.24\linewidth}
        \includegraphics[width=1\textwidth]{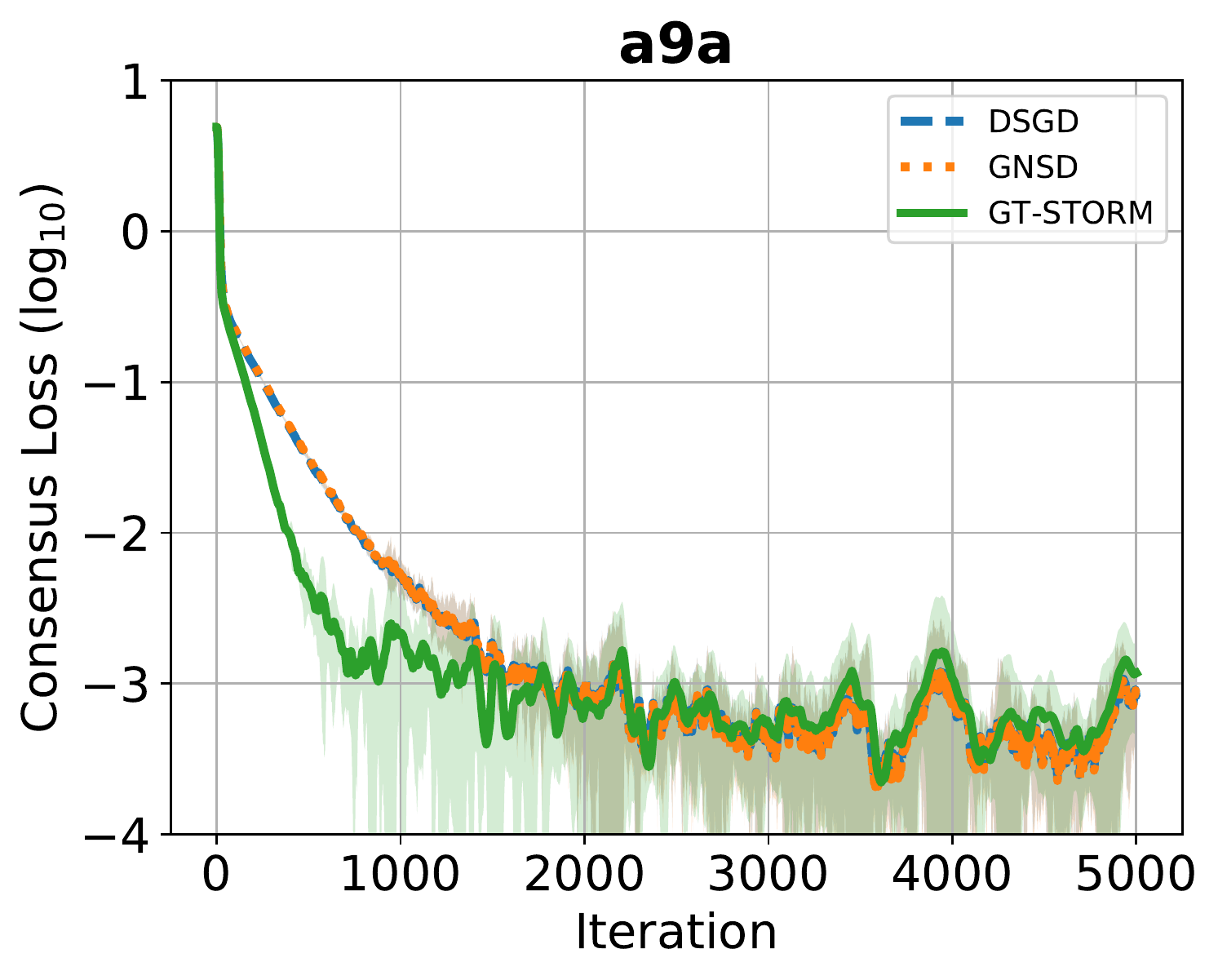}
        \caption{Non-convex logistic regression on LibSVM: a9a.} \label{fig_a}
    \end{minipage}%
    \hspace{0.005\linewidth}
    \begin{minipage}[t]{0.24\linewidth}
        \centering
        \includegraphics[width=1\textwidth]{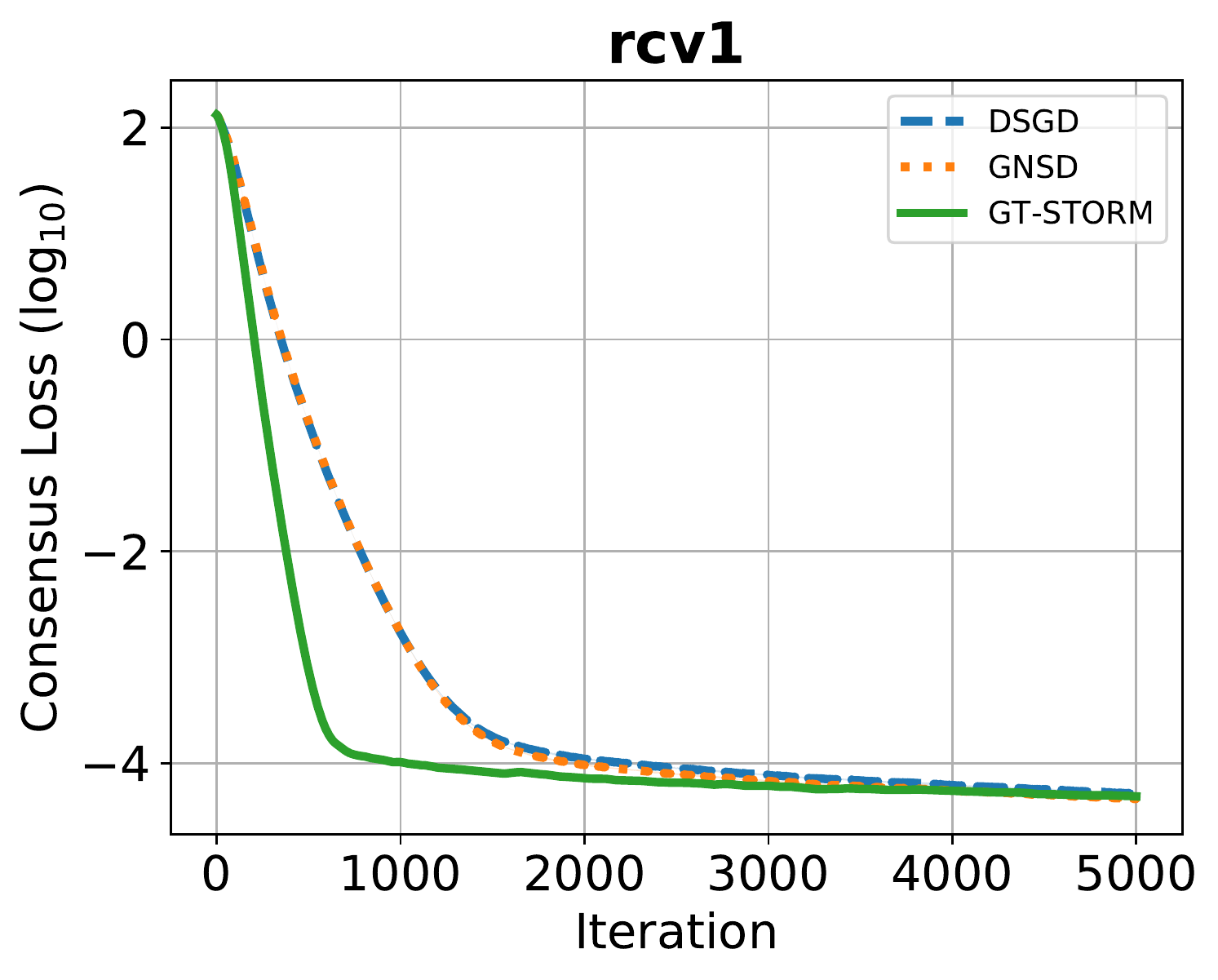}
        \caption{Non-convex logistic regression on LibSVM: ijcnn1.} \label{fig_b}
    \end{minipage}%
    \hspace{0.005\textwidth}
    \begin{minipage}[t]{0.24\linewidth}
        \includegraphics[width=1\textwidth]{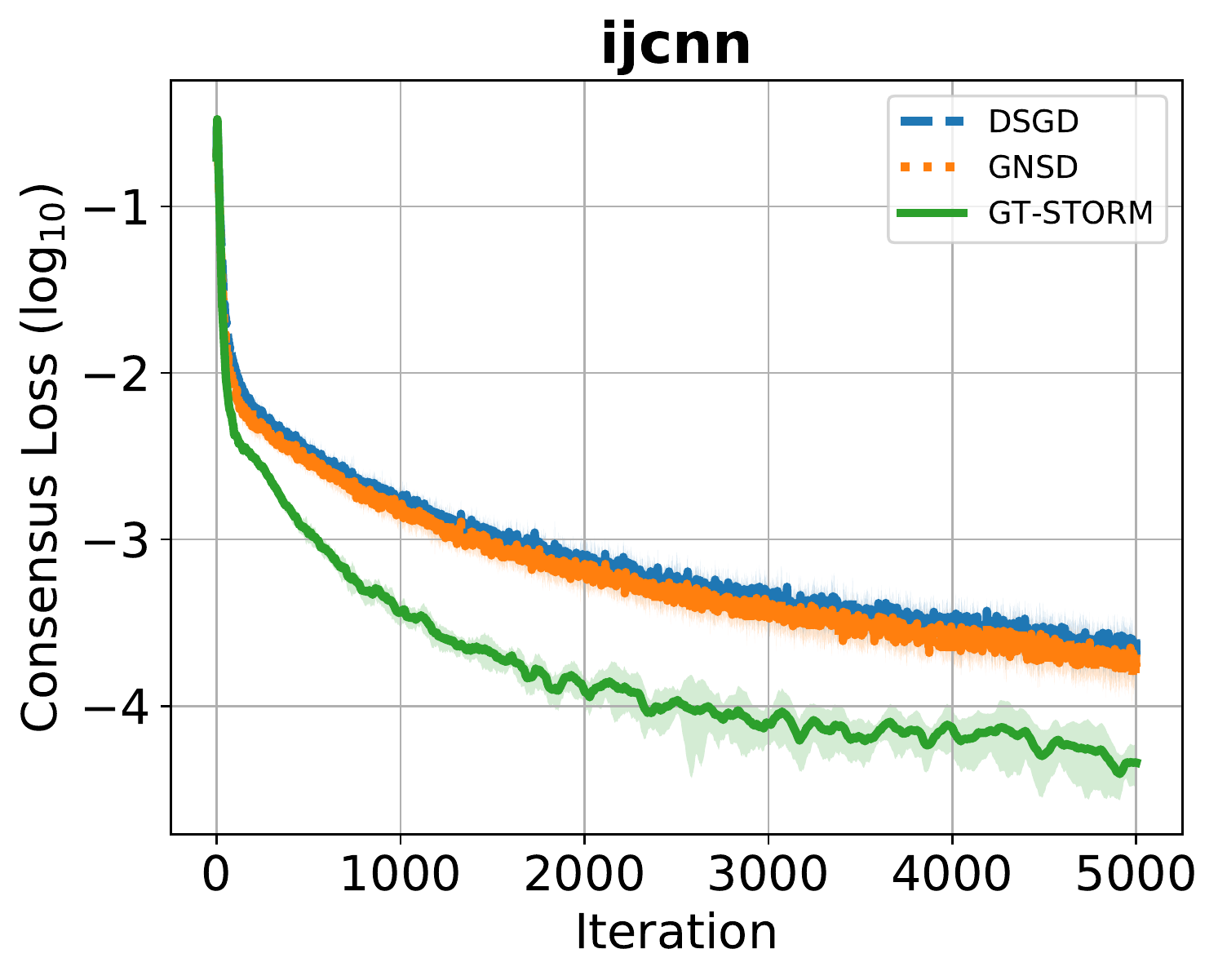}
        \caption{Non-convex logistic regression on LibSVM: rcv1.}\label{fig_c}
    \end{minipage}%
    \hspace{0.005\linewidth} 
    \begin{minipage}[t]{0.24\linewidth}
        \includegraphics[width=1\textwidth]{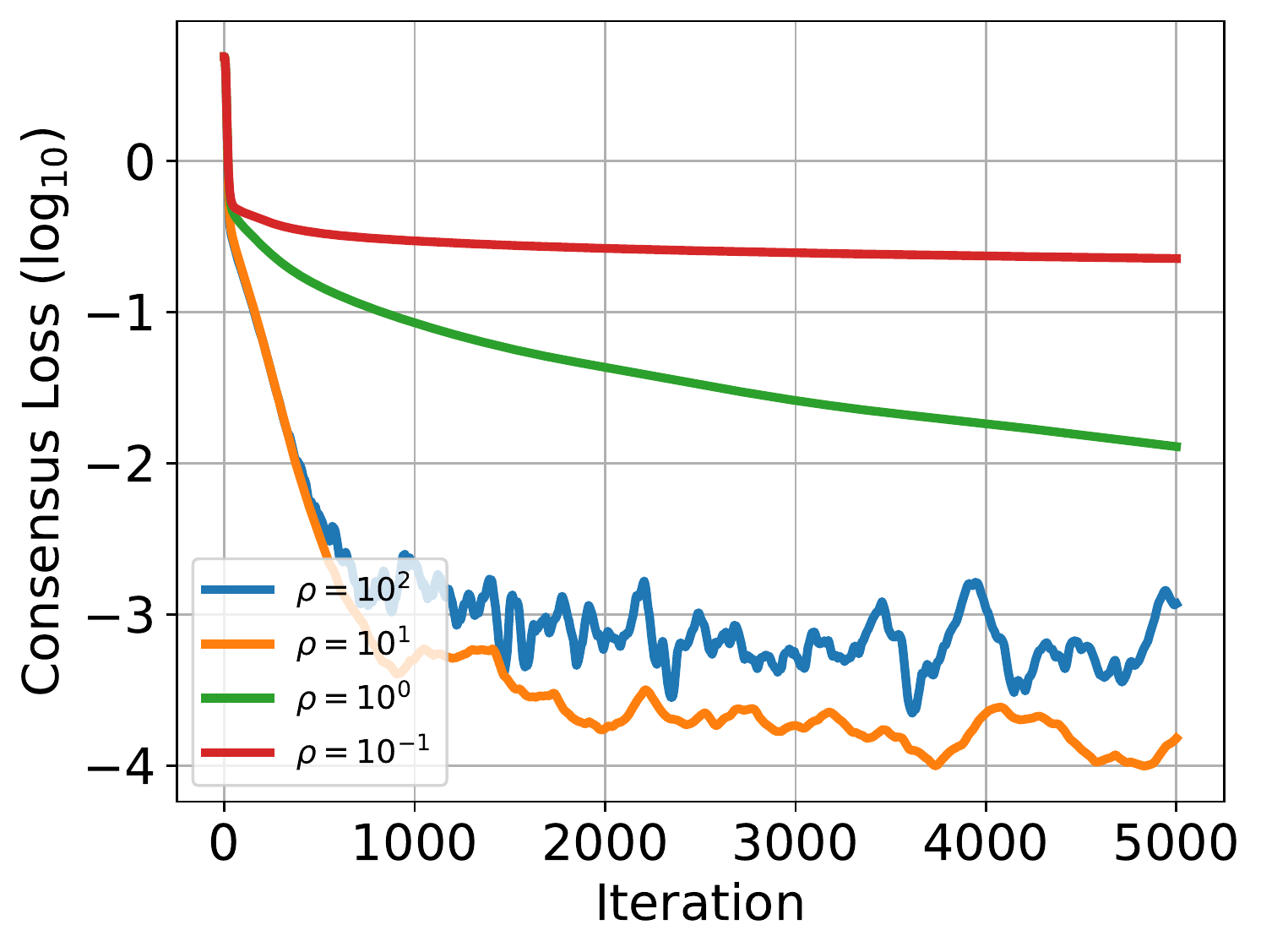}
        \caption{Non-convex logistic regression: The effect of $\rho$.}\label{fig_d}
    \end{minipage}%
\vspace{-.1in}
\end{figure*}

\smallskip
{\em 1-c) \underline{Results:}} 
We first compare the convergence rates of the algorithms. 
We adopt the consensus loss defined in the left-hand-side (LHS) of (\ref{Eq: FOSP_network}) as the criterion.
After tuning, the best initial step-sizes are $0.1,$ $0.5$ and $0.2$ for $a9a$, $ijcnn1$ and $rcv1.binary,$ respectively.
The results are shown in Figs.~\ref{fig_a}--\ref{fig_c}.
It can be seen that our algorithm has a better performance: for $a9a$ and $rcv1.binary$ datasets, all algorithms reach almost the same accuracy but our algorithm has a faster speed; for $ijcnn1$ dataset, our algorithm outperforms other methods both in the speed and accuracy.

\begin{figure*}[t!]
 \begin{minipage}[t]{0.24\linewidth}
        \includegraphics[width=1\textwidth]{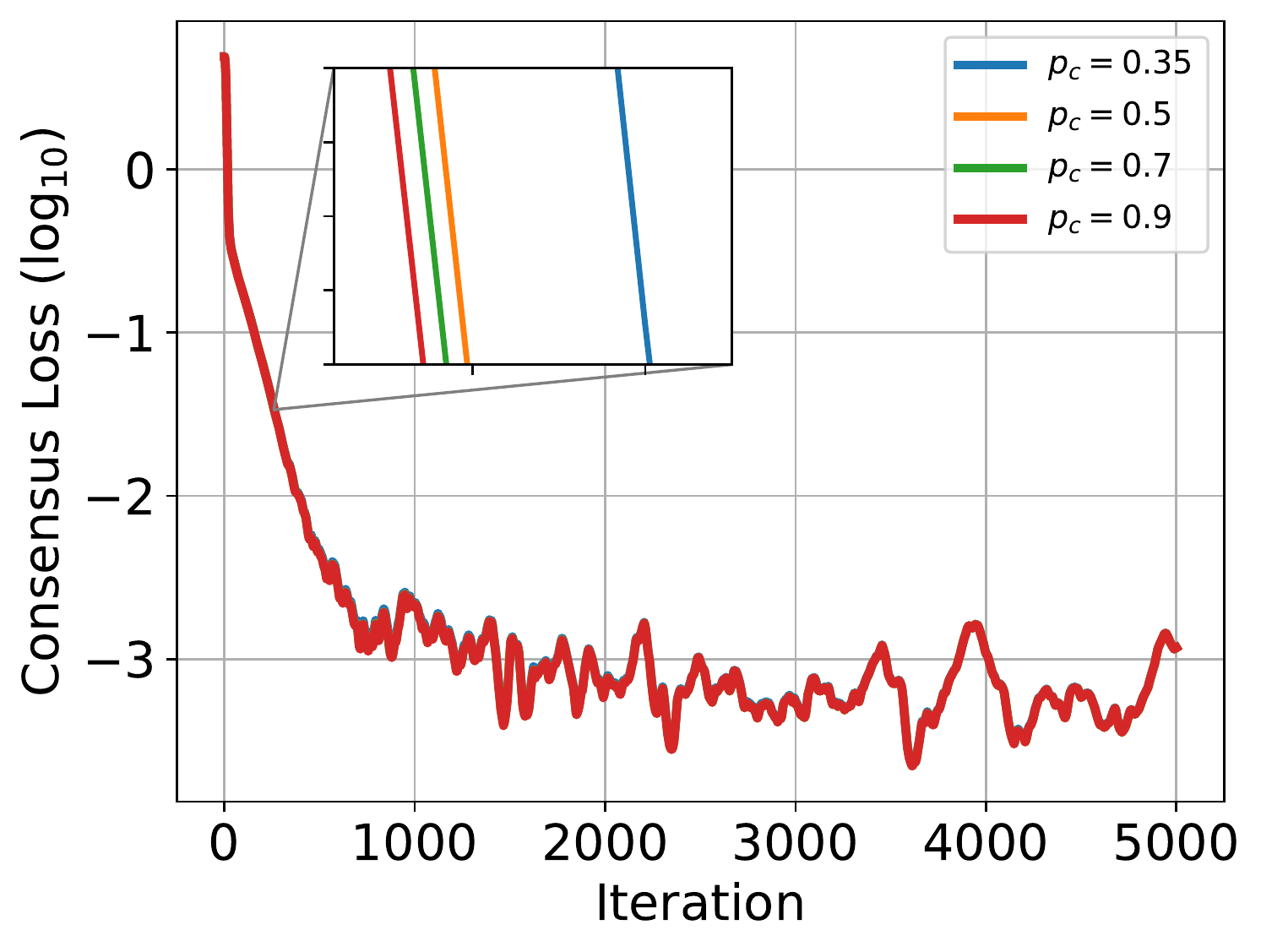}
        \caption{Non-convex logistic regression: The effect of $p_c$.} \label{fig_e}
    \end{minipage}%
    \hspace{0.005\linewidth}
    \begin{minipage}[t]{0.24\linewidth}
        \centering
        \includegraphics[width=1\textwidth]{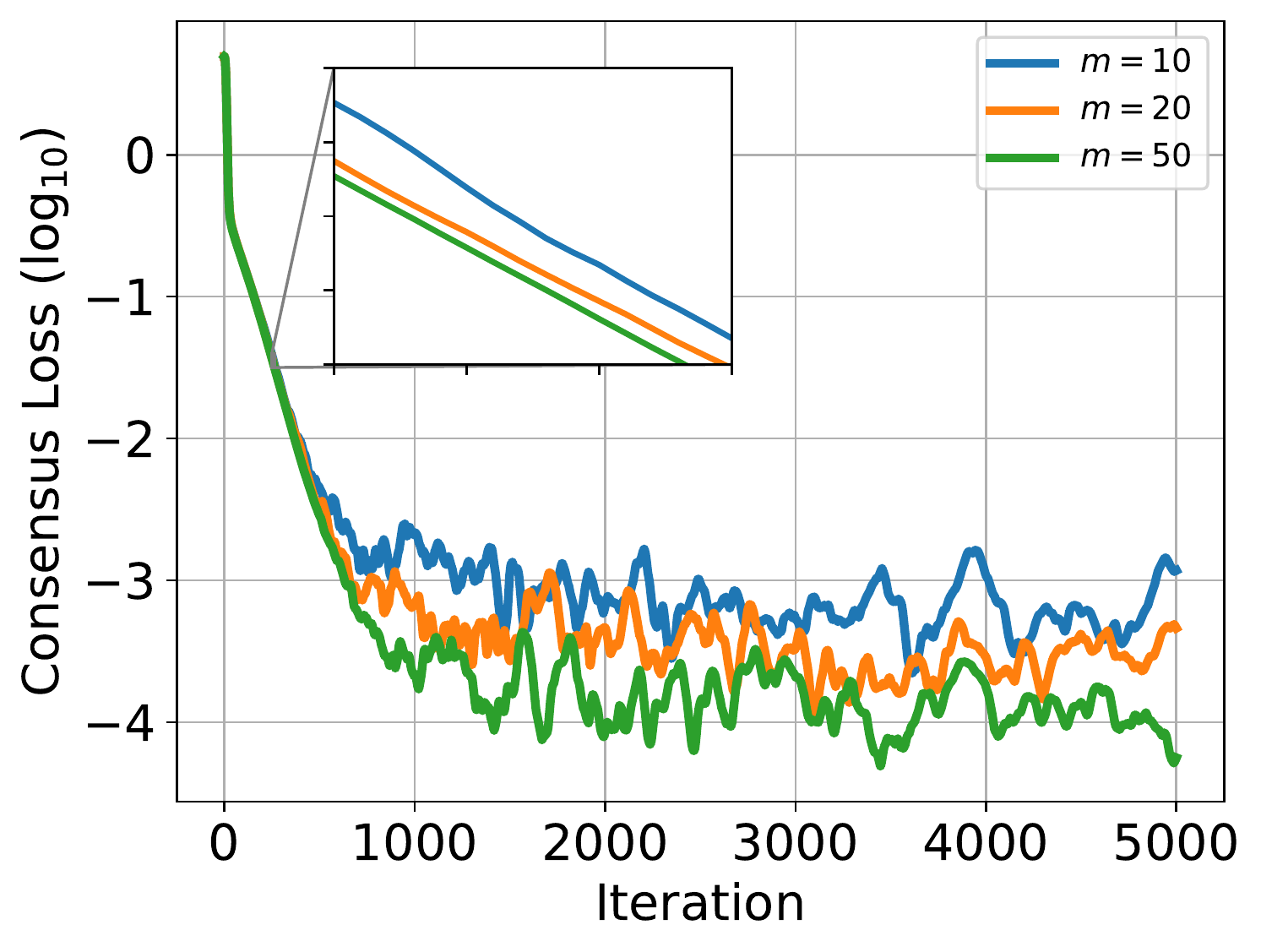}
        \caption{Non-convex logistic regression: The effect of $m$.} \label{fig_f}
    \end{minipage}%
    \hspace{0.005\textwidth}
    \begin{minipage}[t]{0.24\linewidth}
        \includegraphics[width=1\textwidth]{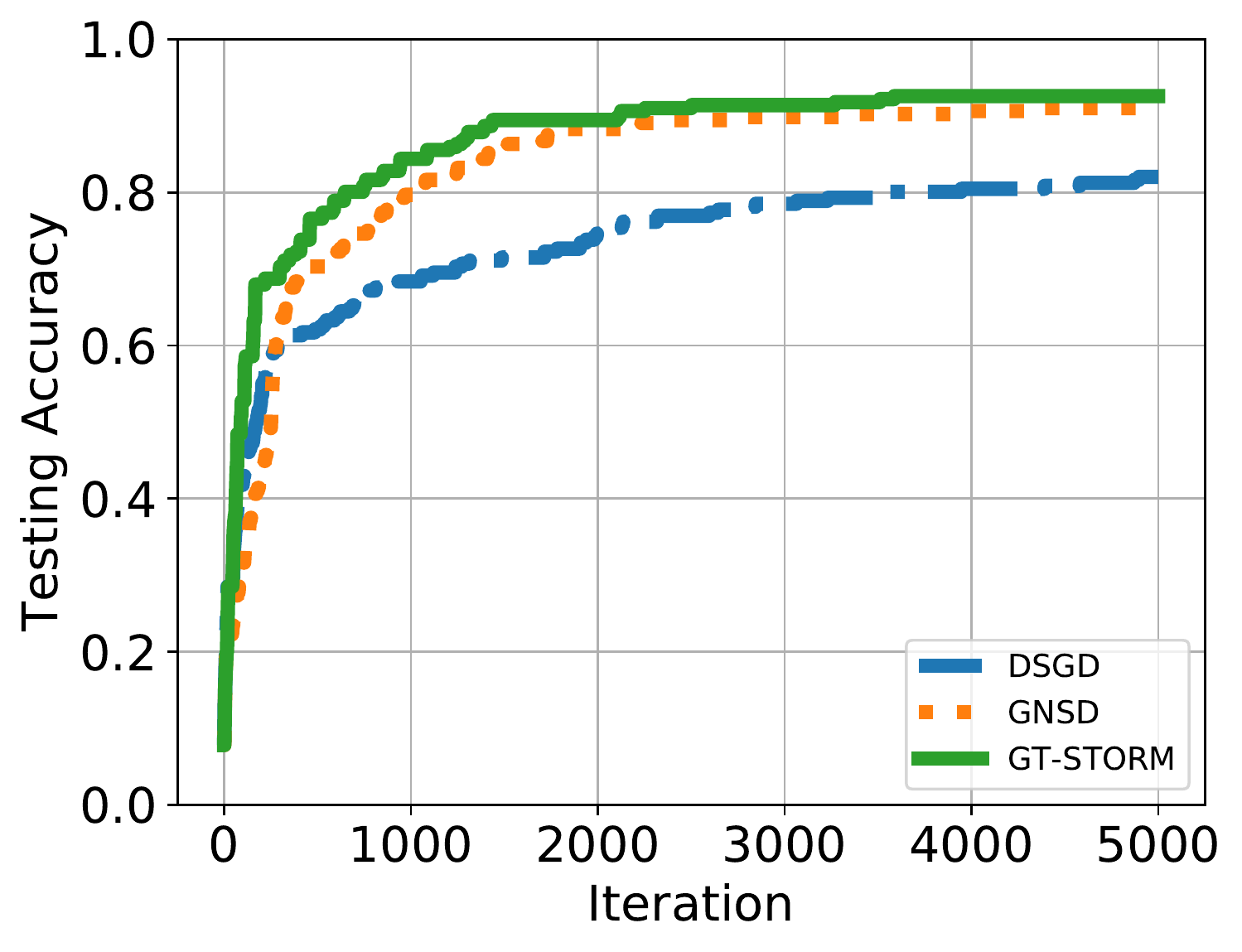}
        \caption{CNN experimental results on MNIST dataset.}\label{fig_g}
    \end{minipage}%
    \hspace{0.005\linewidth} 
    \begin{minipage}[t]{0.24\linewidth}
        \includegraphics[width=1\textwidth]{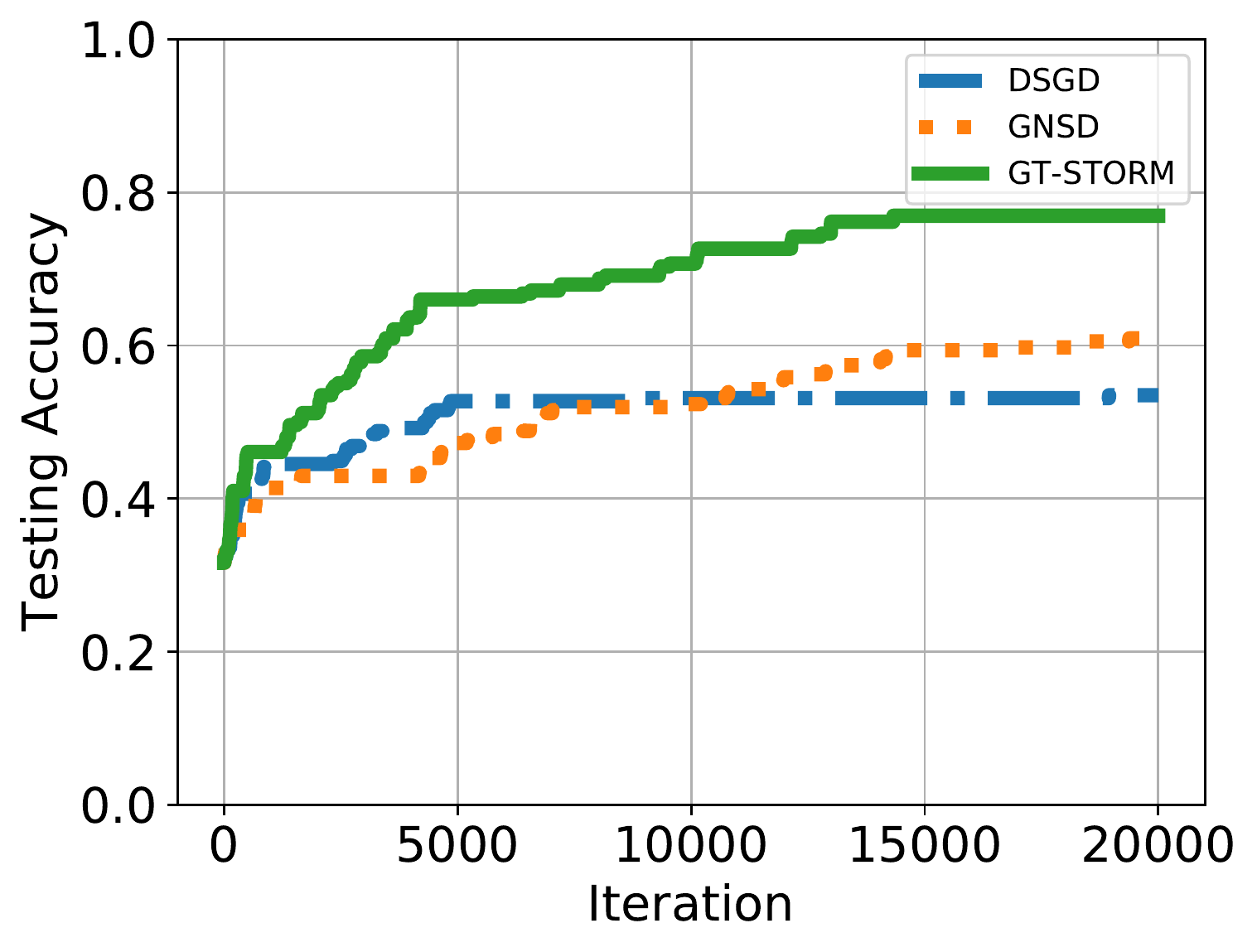}
        \caption{CNN experimental results on CIFAR-10 dataset.}\label{fig_h}
    \end{minipage}%
\vspace{-.15in}
\end{figure*}

Next, we examine the effect of the parameter $\rho$ on our algorithm. 
We focus on the $a9a$ dataset and fix the initial step-size as $\eta_0\!=\!0.1$.
We choose $\rho$ from $\{10^{-1},10^0, 10^1,10^2\}.$ 
Note that $\rho = 10^2$ is corresponding to the case $\rho \!=\! 1/\eta_0^2.$ 
The results are shown in Fig.~\ref{fig_d}.
It can be seen that the case $\rho \!=\! 10^1$ has the best performance, which is followed by the case $\rho = 10^2.$ 
Also, as $\rho$ decreases, the convergence speed becomes slower (see the cases $\rho \!=\! 10^{-1}$ and $10^0$).


In addition, we examine the effect of the network topology.
We first fix the number of workers as $m = 10$ and change the the edge connectivity probability $p_c$ from $0.35$ to $0.9.$
Note that with a smaller $p_c,$ the network becomes sparser.
We set $\eta_0 = 0.1$ and $\rho = 10^2.$
The results are shown in Fig.~\ref{fig_e}.
Under different $p_c$-values, our algorithm has a similar performance in terms of convergence speed and accuracy.
But with a larger $p_c$-values i.e., a denser network, the convergence speed slightly increases (see the zoom-in view in Fig.~\ref{fig_f}.
Then, we fix the the edge connectivity probability $p_c = 0.5$ but change the number of workers $m$ from $10$ to $50.$
We show the results in Fig.~\ref{fig_f}.
It can be seen that with more workers, the algorithm converges faster and reaches a better accuracy.

\smallskip
{\bf 2) Convolutional neural networks}\label{Section: Experiment_CNN}
We use all three algorithms to train a convolutional neural network (CNN) model for image classification on MNIST and CIFAR-10 datasets.
We adopt the same network topology as in the previous experiment. 
We use a non-identically distributed data partition strategy: the $i$th machine can access the data with the $i$th label.
We fix the initial step-size as $\eta_0 = 0.01$ for all three algorithms and the remaining settings are the same as in the previous experiment.

{\em 2-a) \underline{Learning Models:}} 
For MNIST, the adopted CNN model has two convolutional layers (first of size $1 \times 16 \times 5$ and then of size $16 \times 32 \times 5$), each of which is followed by a max-pooling layer with size $2\times 2$, and then a fully connected layer. 
The ReLU activation is used for the two convolutional layers and the ``softmax'' activation is applied at the output layer. 
The batch size is 64 for the CNN training on MNIST.
For CIFAR-10, we apply the CNN model with two convolutional layers (first of size $3 \times 6 \times 5$ and then of size $6 \times 16 \times 5$).
Each of the convolutional layers is followed by a max-pooling layer of size $2\times 2$, and then three fully connected layers.
The ReLU activation is used for the two convolutional layers and the first two fully connected layers, and the ``softmax'' activation is applied at the output layer. 
The batch size is chosen as 128 for the CNN training on CIFAR-10.


{\em 2-b) \underline{Results:}} 
Fig.~\ref{fig_g} illustrates the testing accuracy of different algorithms versus iterations on MNIST and CIFAR-10 datasets. 
It can be seen from Fig.~\ref{fig_g} that on the MNIST dataset, GNSD and GT-STORM have similar performance, but our GT-STORM maintains a faster speed and a better prediction accuracy.
Compared with DSGD, our GT-STORM can gain about $10\%$ more accuracy.
On the CIFAR-10 dataset (see Fig.~\ref{fig_h}), the performances of DSGD and GNSD deteriorate, while GT-STORM can achieve a better accuracy.
Specifically, the accuracy of GT-STORM is around $15\%$ higher than that of GNSD and $25\%$ higher than that of DSGD.

\section{Conclusion}\label{Section: conclusion}

In this paper, we proposed a gradient-tracking-based stochastic recursive momentum (GT-STORM) algorithm for decentralized non-convex optimization, which enjoys low sample, communication, and memory complexities.
Our algorithm fuses the gradient tracking estimator and the variance reduction estimator and has a simple single-loop structure.
Thus, it is more practical compared to existing works (e.g. GT-SAGA/SVRG and D-GET) in the literature.
We have also conducted extensive numerical studies to verify the performance of our method, including non-convex logistic regression and neural networks.
The numerical results show that our method outperforms the state-of-the-art methods when training on the large datasets.
Our results in this work contribute to the increasingly important field of decentralized network training.

\bibliographystyle{IEEEtran}
\bibliography{reference}

\appendix
\section{Addtional Experiment Details}

\begin{wrapfigure}{R}{0.4\columnwidth}
\centering
 \includegraphics[width=1.0\linewidth]{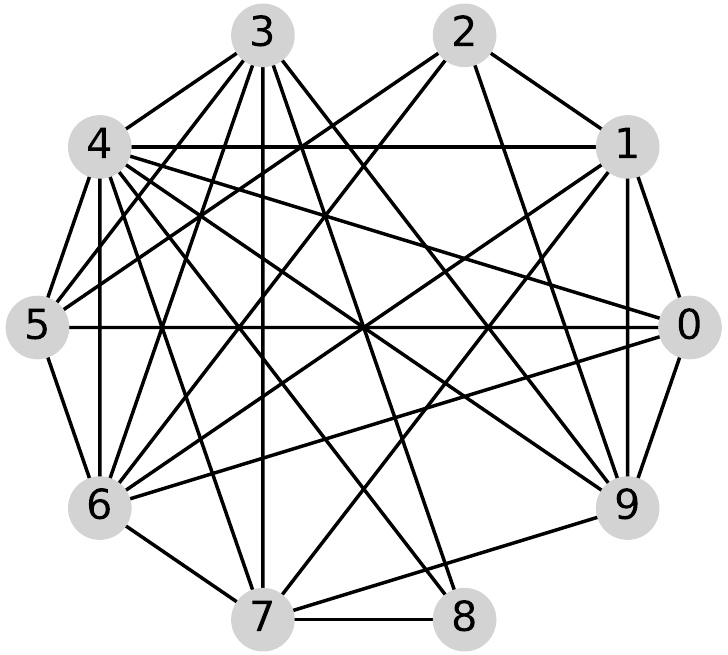}
\vspace{-.1in}
\caption{The network topology generated by the Erd$\ddot{\text{o}}$s-R$\grave{\text{e}}$nyi graph.}
\label{fig: Sys}
\vspace{-.3in}
\end{wrapfigure}

In our simulation, the communication graph $\mathcal{G}$ is generated by the Erd$\ddot{\text{o}}$s-R$\grave{\text{e}}$nyi graph with different edge connectivity probability $p_c$ and nodes number $m$.
We set $m=10$ and $p_c =0.5$. The generated graph is shown in Figure \ref{fig: Sys}.

\subsection{Nonconvex Logistic Regression}

\begin{figure*}[!ht]
\begin{minipage}{0.24\textwidth}
    \centering
    \begin{tabular}{c}  
    \includegraphics[width = 1\textwidth]{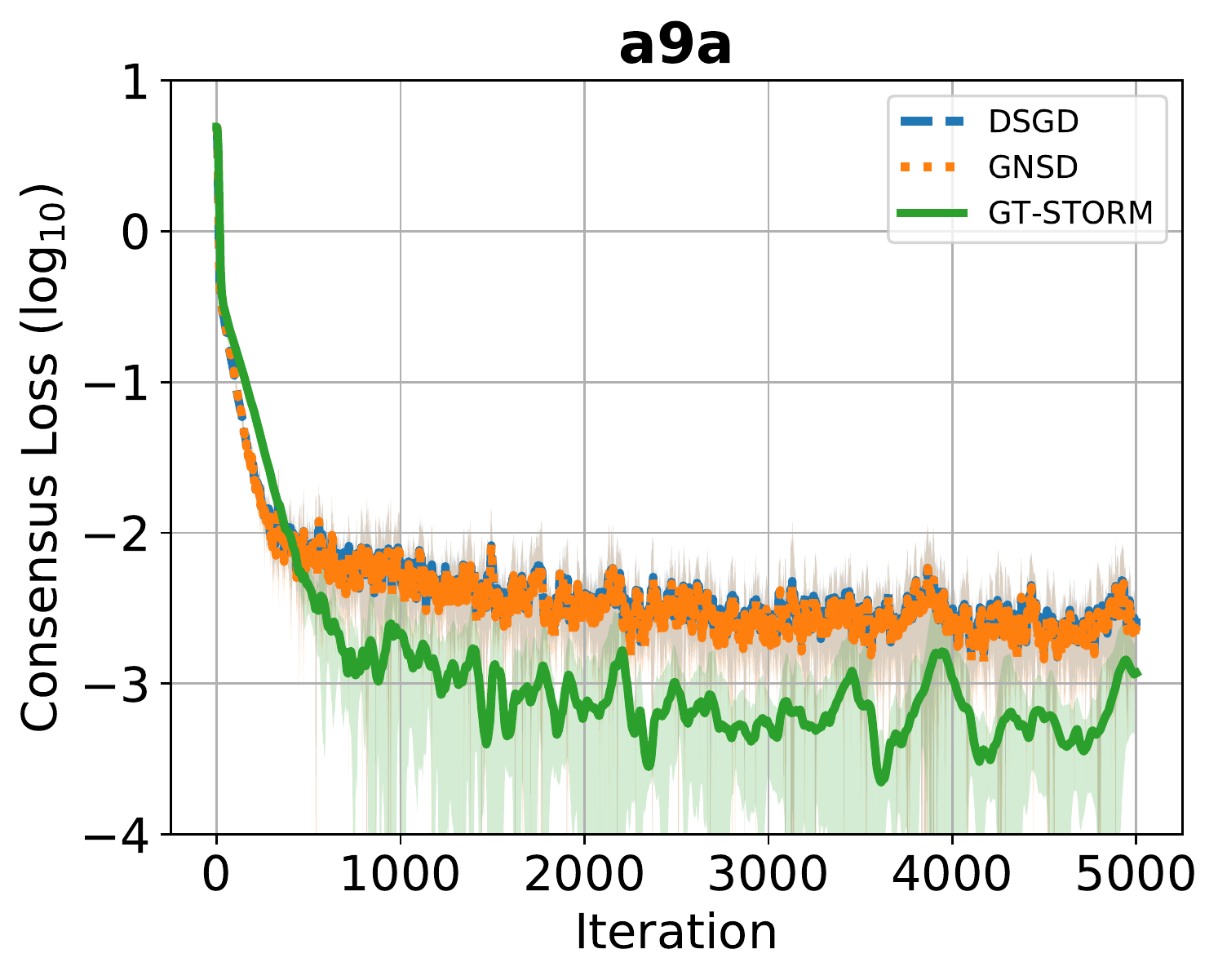}
\end{tabular}
    \end{minipage}
    \hspace{0.1in}
    \begin{minipage}{0.24\textwidth}
    \centering
        \begin{tabular}{c}  
        \includegraphics[width = 1\textwidth]{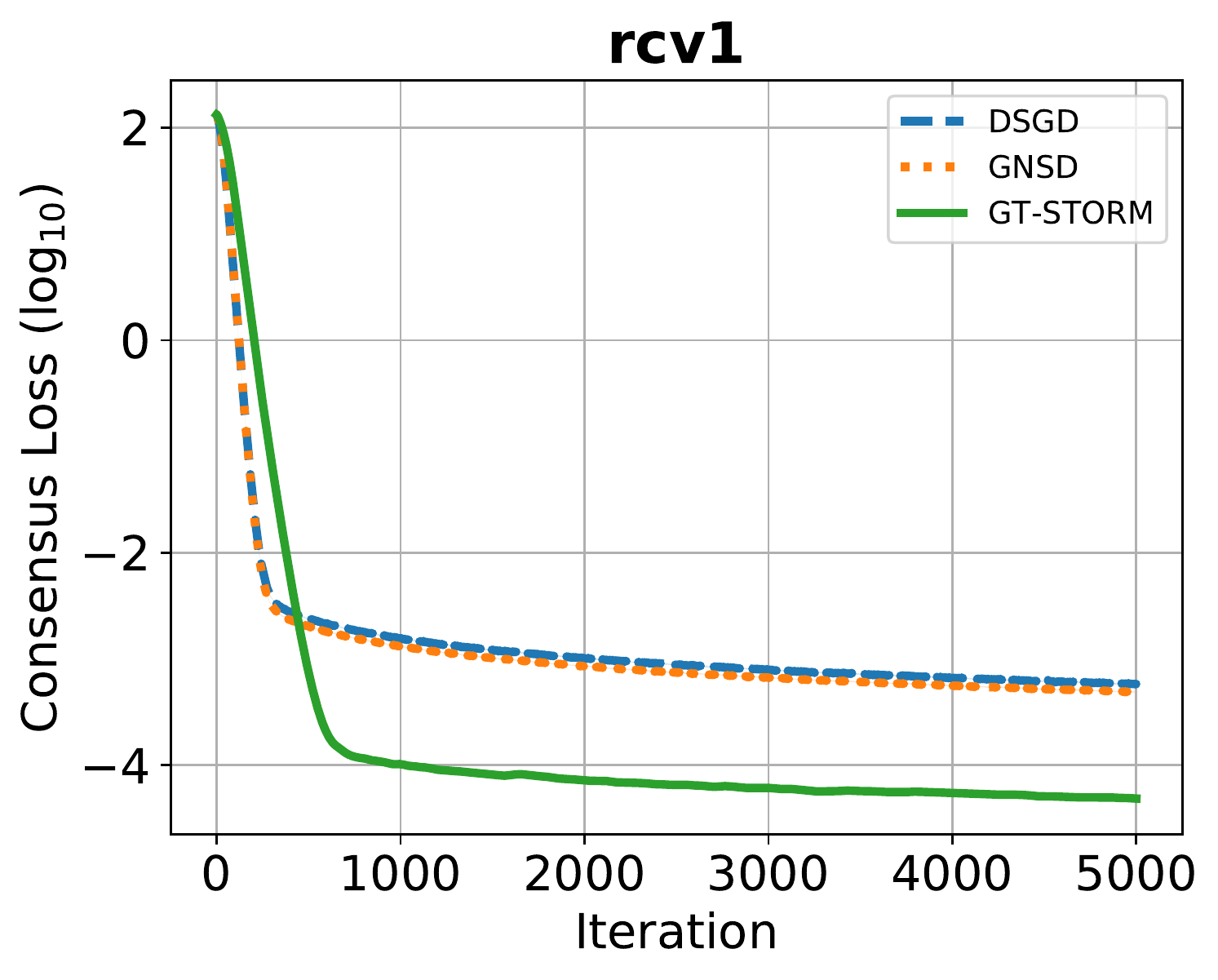}
        \end{tabular}
    \end{minipage}
    \hspace{0.1in}
    \begin{minipage}{0.24\textwidth}
    \centering
        \begin{tabular}{c}  
        \includegraphics[width = 1\textwidth]{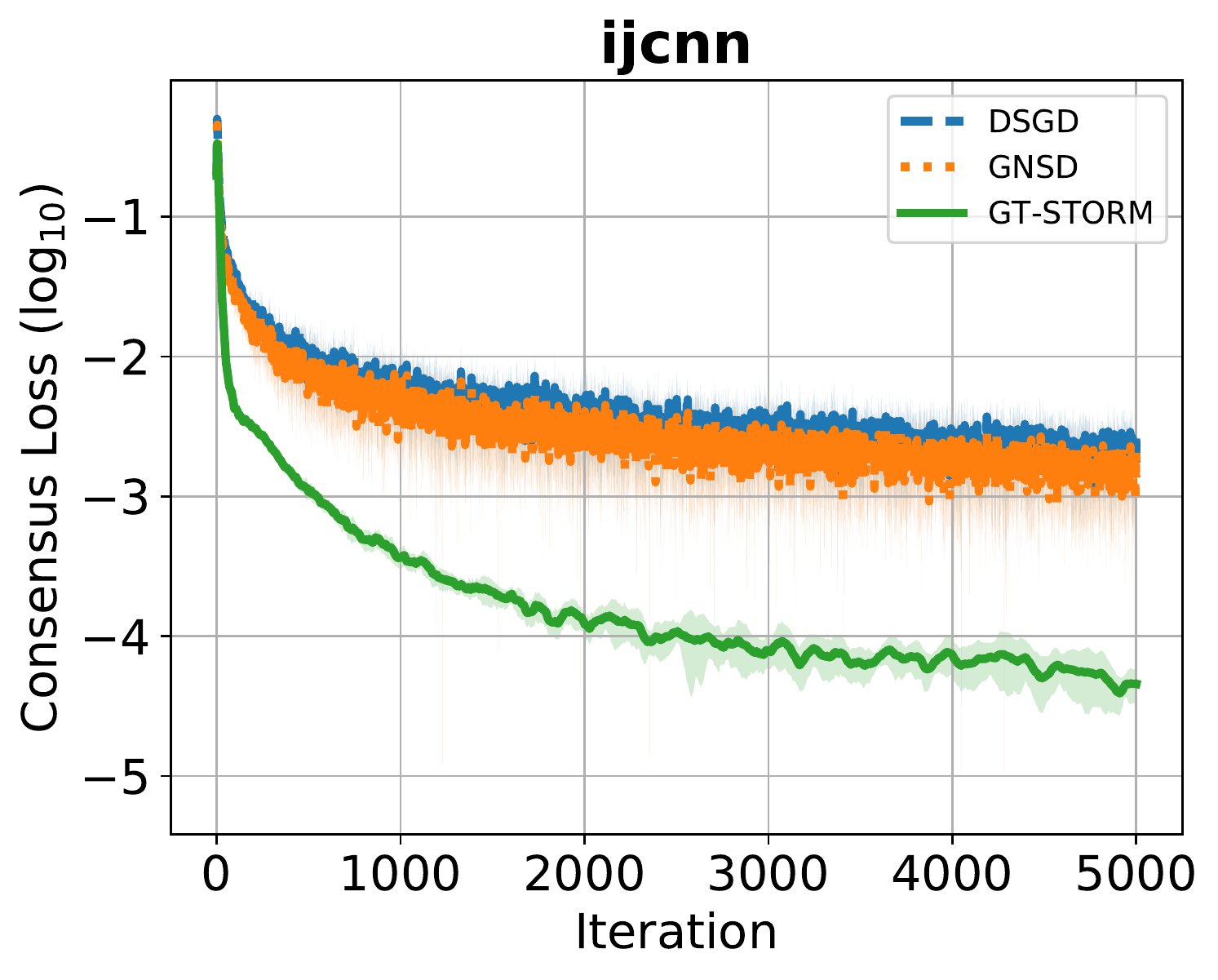}
        \end{tabular}
    \end{minipage}
    \vspace{-.1in}
\caption{Experimental results for nonconvex logistic regression on LibSVM datasets: the three algorithms adopt a diminishing step-size $\eta_t = \eta_0/\sqrt[3]{1+0.1\times t};$ the curves are averaged over 10 random trials and the shaded regions represent the standard deviation.}\label{Fig: lr_simu1_}
\end{figure*}

In Section 4.1, we set the step-size as $\eta_t = \eta_0/\sqrt{1+0.1\times t}$ for DSGD and GNSD, while $\eta_t = \eta_0/\sqrt[3]{1+0.1\times t}$ for GT-STORM.
It can be noted that the step-size adopted for GT-STORM is diminishing slower than those for DSGD and GNSD, though the choices are following the theoretical results.
Thus, here we apply the step-size as $\eta_t = \eta_0/\sqrt[3]{1+0.1\times t}$ for all the three algorithms.
We tune the initial step-size $\eta_0$ by searching the grid $\{0.01, 0.02, 0.05 , 0.1, 0.2,  0.5, 1.0\}.$ 
After tuning, the best initial step-sizes are $0.1,$ $0.5$ and $0.2$ for $a9a$, $ijcnn1$ and $rcv1.binary,$ respectively.
We show the results in Figure \ref{Fig: lr_simu1_}.
It can be seen that with a larger step-size, though the convergence is faster for DSGD and GNSD at the beginning, the accuracy is unsatisfactory (e.g. $a9a$ and $rcv1.binary$).
Also, with the same step-size, our algorithm performs much better than the other two.

\subsection{Convolutional Neural Networks}

\begin{figure*}[!ht]
    \centering
    \begin{tabular}{cccc}  
    \includegraphics[width = 0.24\textwidth]{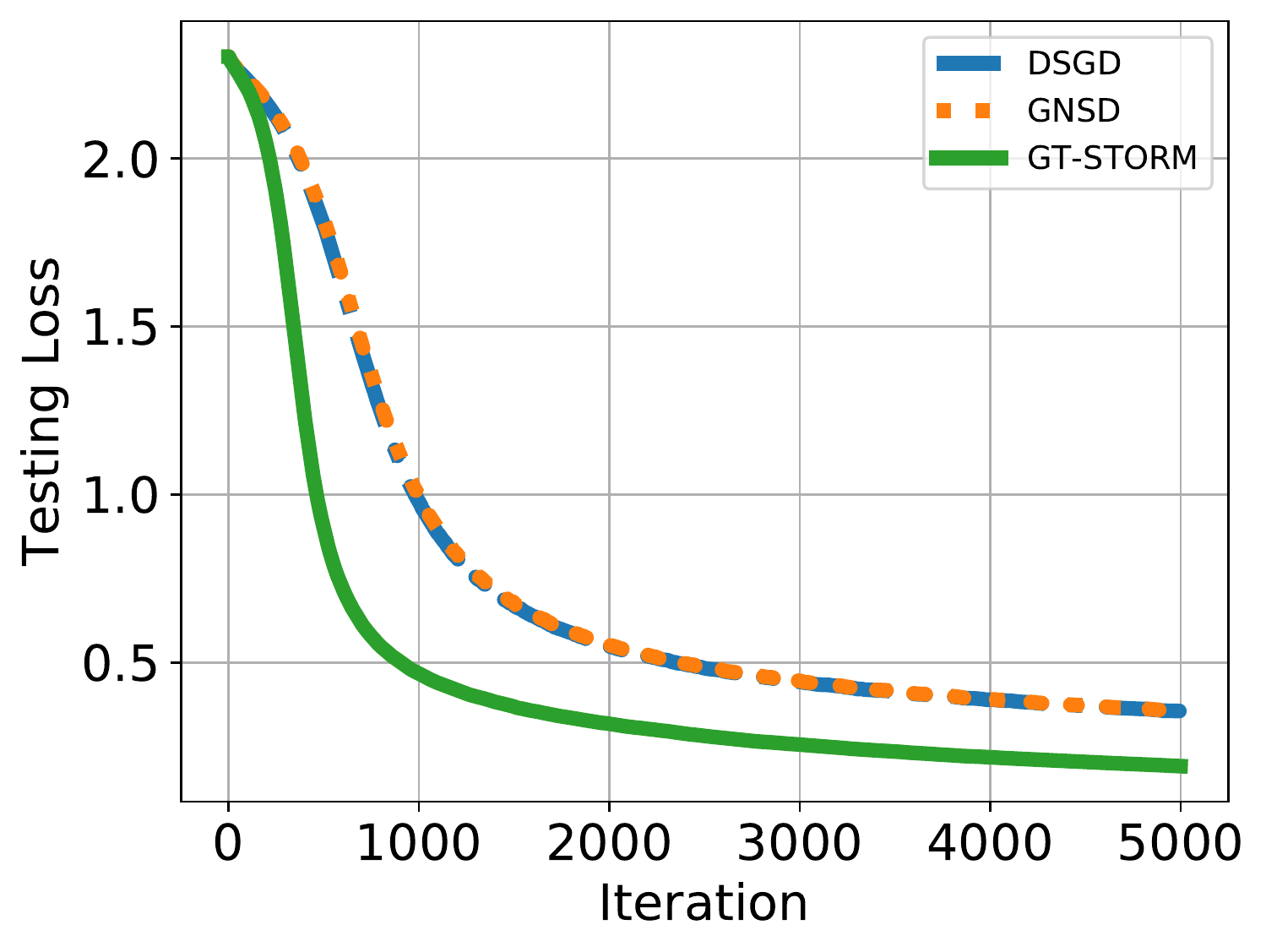}&
    \includegraphics[width = 0.24\textwidth]{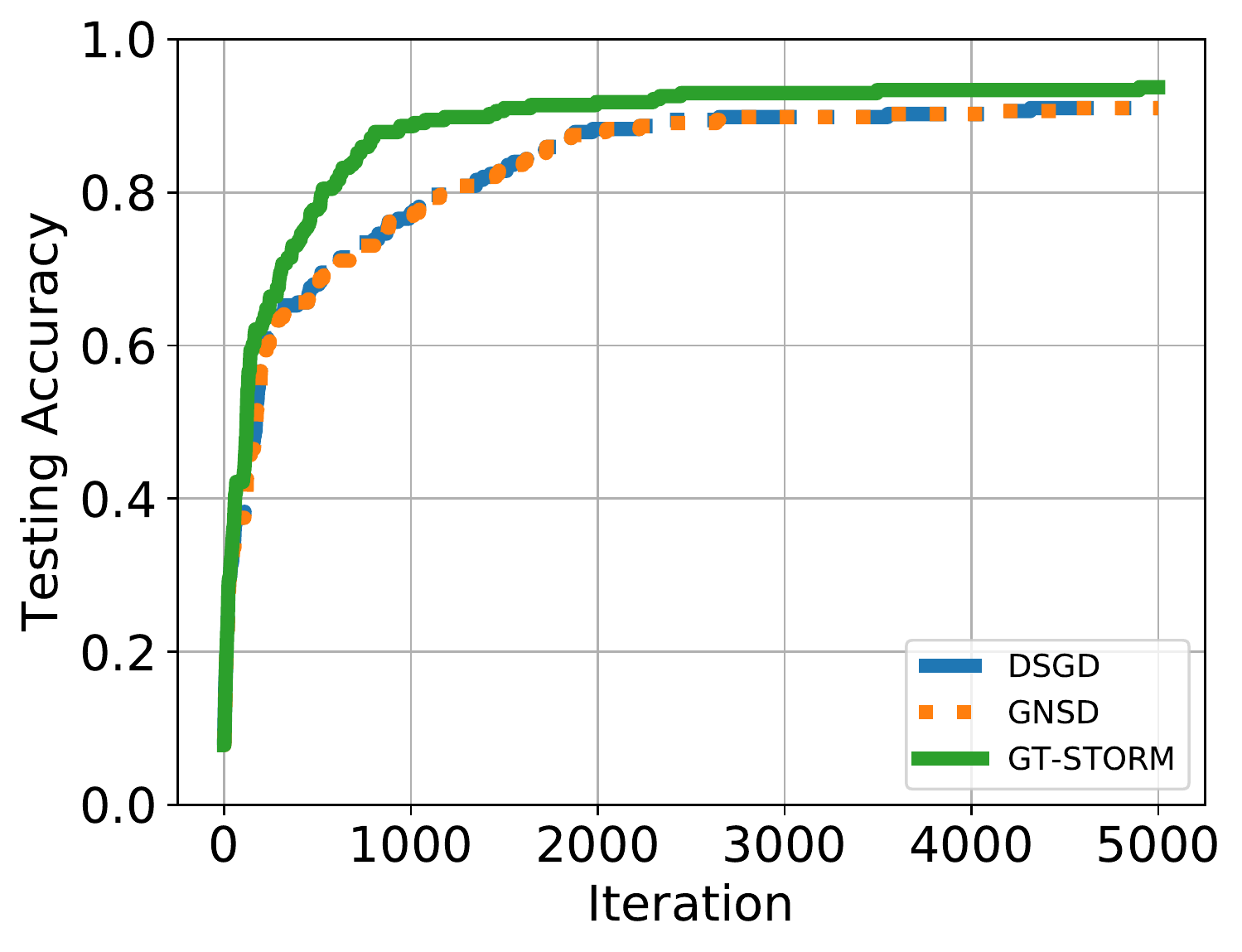}&
    \includegraphics[width = 0.24\textwidth]{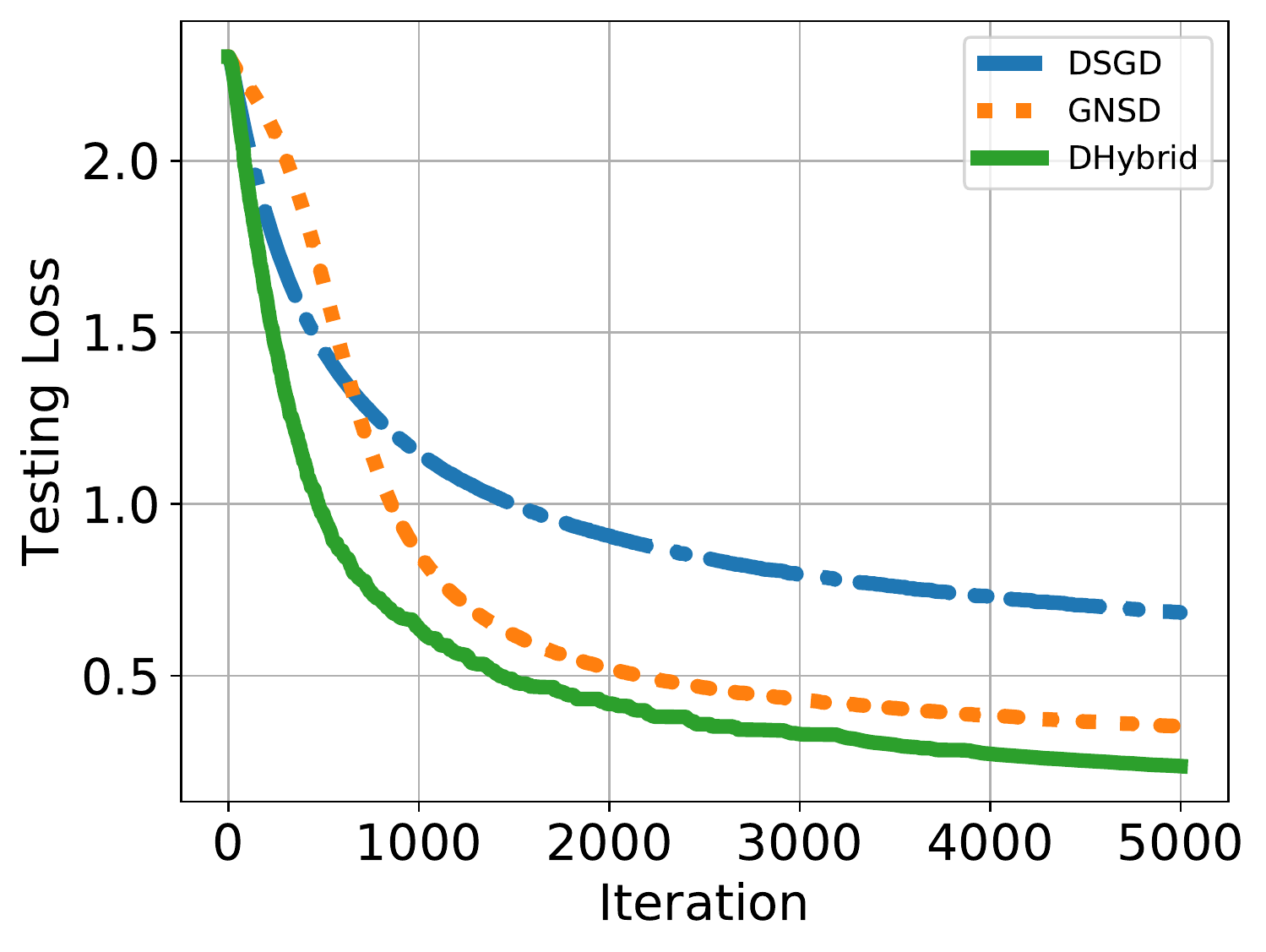}&
    \includegraphics[width = 0.24\textwidth]{dnn_mnist2_accy.pdf}\\ 
         \multicolumn{2}{c}{(a) I.D. data partition} &\multicolumn{2}{c}{(b) N.D. data partition}\\ 
\end{tabular}
    \caption{Experimental results for CNN on MNIST dataset.}\label{Fig: cnn_mnist_appdx}
\end{figure*}

\begin{figure*}[!ht]
    \centering
    \begin{tabular}{cccc}  
    \includegraphics[width = 0.24\textwidth]{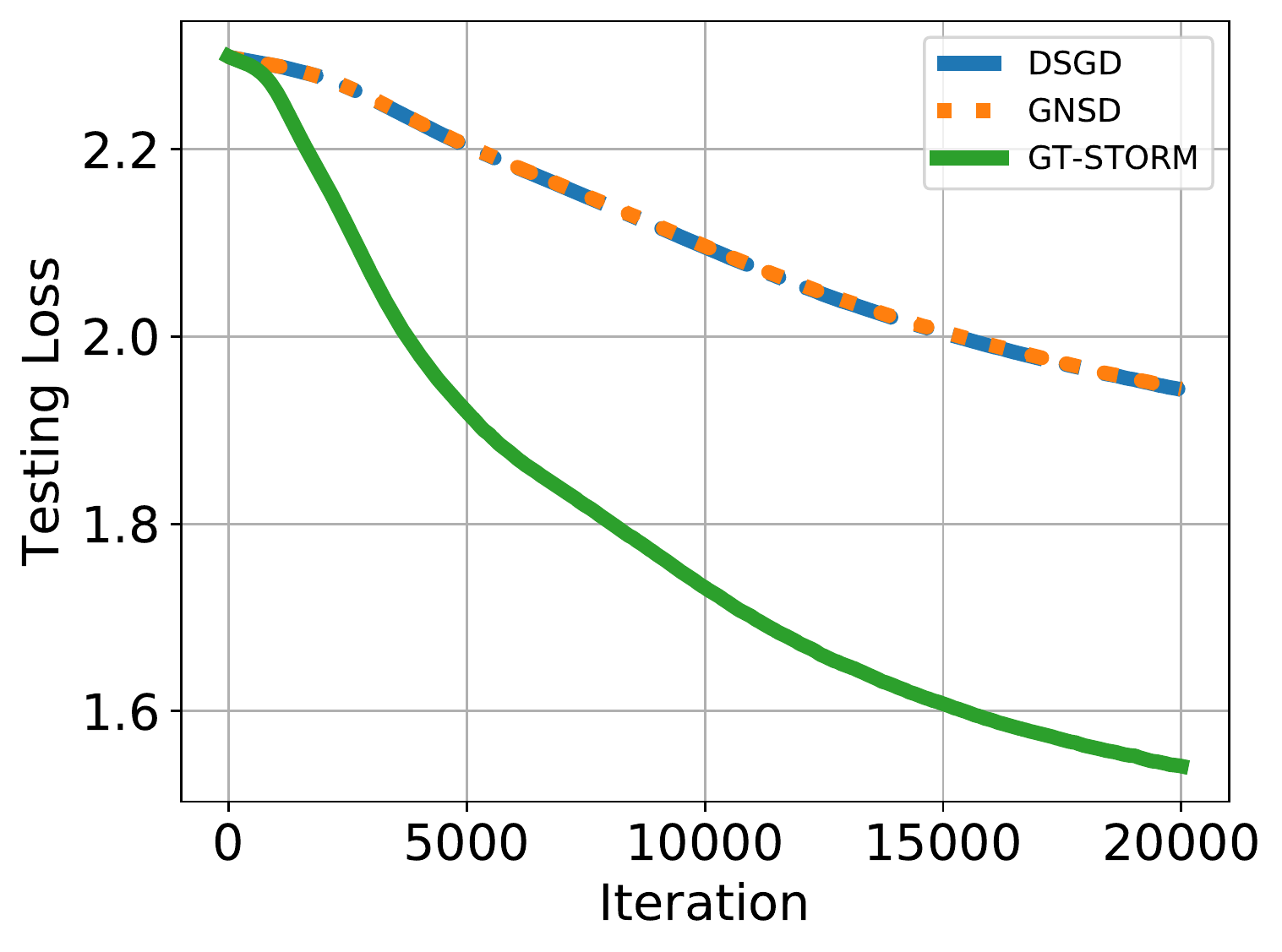}&
    \includegraphics[width = 0.24\textwidth]{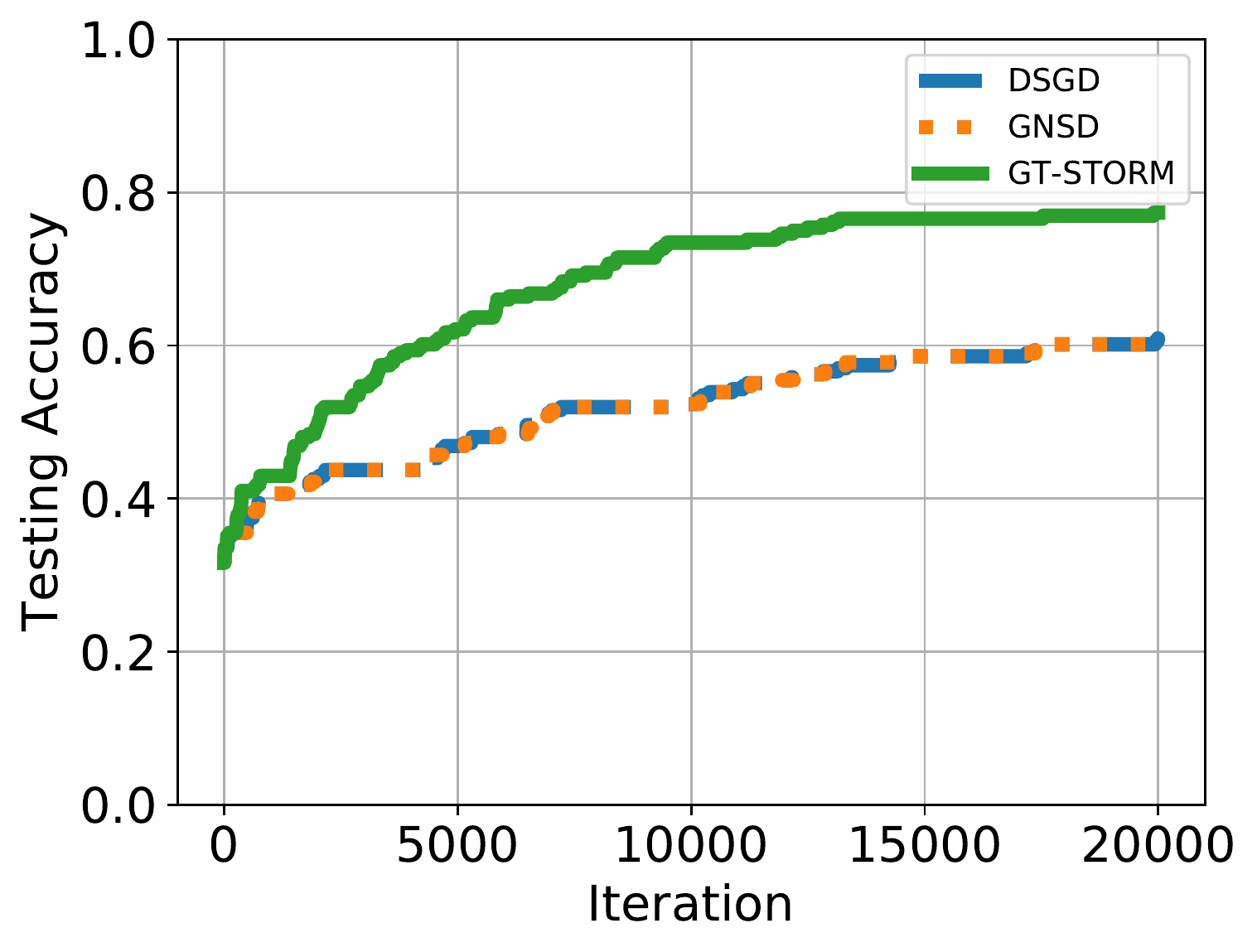}&
    \includegraphics[width = 0.24\textwidth]{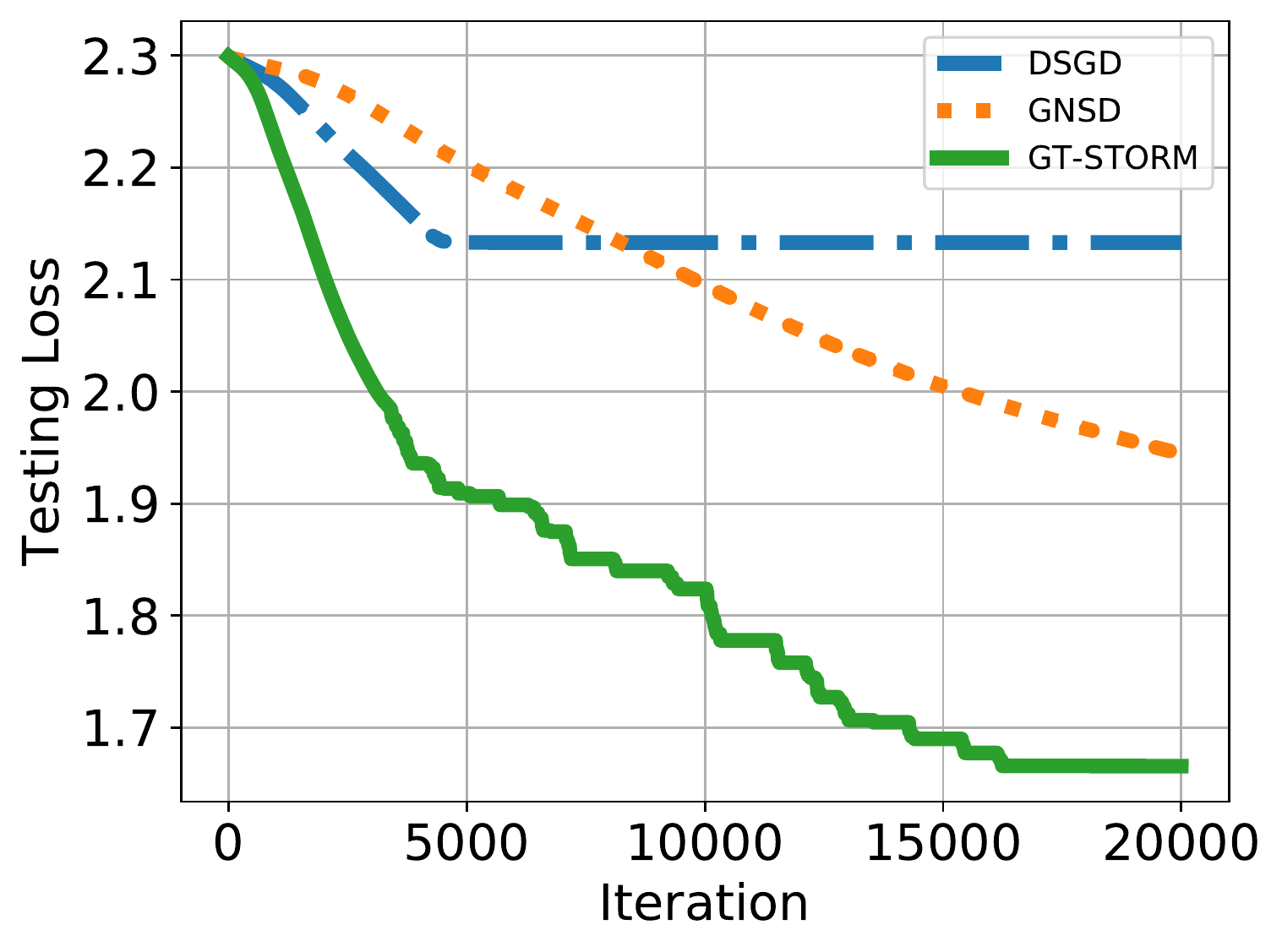}&
    \includegraphics[width = 0.24\textwidth]{dnn_cifar2_accy.pdf}\\ 
         \multicolumn{2}{c}{(a) I.D. data partition} & \multicolumn{2}{c}{(b) N.D. data partition}\\ 
\end{tabular}
    \caption{Experimental results for CNN on CIFAR-10 dataset.}\label{Fig: cnn_cifar_appdx}
\end{figure*}

Here we show the testing loss and accuracy for the CNN models on the MNIST and CIFAR-10 datasets in Figure \ref{Fig: cnn_mnist_appdx}-\ref{Fig: cnn_cifar_appdx}. 
In all experiment results, our algorithm has a better performance: a higher accuracy and a smaller loss.
The final testing accuracy results are summarized in Table \ref{Table: Testing Accuracy}.

\begin{table}[!ht]
\caption{The results of testing accuracy with the CNN models trained by different algorithms.}\label{Table: Testing Accuracy}
\vspace{-.in}
\begin{center}
\begin{tabular}{ccccc}
\hline
\hline
\multicolumn{2}{c}{Dataset} & DSGD & GNSD & GT-STORM \\
\hline
\multirow{2}{*}{MNIST}  & I.D. & 0.9102 & 0.9102 & 0.9375\\
                        & N.D. & 0.8203 & 0.9102 & 0.9257\\
                        \hline
\multirow{2}{*}{CIFAR-10}  & I.D. & 0.6093 & 0.6016 & 0.7734\\
                        & N.D. & 0.5352 & 0.6133 & 0.7695 \\
\hline
\hline
\end{tabular}
\end{center}
\vspace{-.2in}
\end{table}

\section{Proof of Main Results}

Due to space limitation, we provide key proof steps of the key lemmas and theorems in this appendix.
We refer readers to~\cite{Zhang20:GT-STORM_TR} for the complete proofs.

\subsection{Proof for Lemma 1}

\begin{proof}
Recall that $\bar{\v}_{t} = \beta_t (\bar{\v}_{t-1} +\bar{\w}_t) + (1-\beta_t)\bar{\u}_t$, then
\begin{align}
& 
\|\bv_t \!-\! \bg_t \|^2 
= 
\|\beta_t (\bv_{t\!-\!1} +\bar{\w}_t) \!+\! (1\!-\!\beta_t)\bar{\u}_t\! -\! \bg_t\|^2 \nonumber\\
= 
&
\|\beta_t (\bv_{t\!-\!1}\! -\! \bg_{t\!-\!1}) \!+\!\beta_t(\bar{\w}_t \!-\!\bg_{t} \!+\! \bg_{t\!-\!1}) \!+\! (1\!-\!\beta_t)(\bar{\u}_t \!-\! \bg_{t})\|^2 \notag\\
= 
&
\beta_t^2 \|\bv_{t\!-\!1} \!-\! \bg_{t\!-\!1}\|^2 \!+\! \|\beta_t(\bar{\w}_t \!-\!\bg_{t} \!+\! \bg_{t\!-\!1}) \!+\! (1\!-\!\beta_t)(\bar{\u}_t \!-\! \bg_{t})\|^2 \notag\\
& 
\!+\! 2\langle \bv_{t\!-\!1} \!-\! \bg_{t\!-\!1},\beta_t(\bar{\w}_t \!-\!\bg_{t} \!+\! \bg_{t\!-\!1}) \!+\! (1\!-\!\beta_t)(\bar{\u}_t \!-\! \bg_{t}) \rangle
\end{align}
Note that $\Eb_{\zeta_t}[\bar{\w}_{t}] = \bg_t - \bg_{t-1}$ and  $\Eb_{\zeta_t}[\bar{\u}_{t}] = \bg_t.$ 
Taking expectation with respect to $\zeta_t,$ we have:
{{\begin{align}
& 
\Eb_{\zeta_t}[\|\bv_t - \bg_t \|^2 ]\notag\\
&
\stackrel{(a)}{=}
\beta_t^2 \|\bv_{t\!-\!1}\! -\! \bg_{t\!-\!1}\|^2\! +\! \Eb_{\zeta_t}[\|\beta_t(\bar{\w}_t\! -\!\bg_{t}\! +\! \bg_{t\!-\!1})\! +\! (1\!-\!\beta_t)(\bar{\u}_t \!-\! \bg_{t})\|^2] \notag\\
&
\stackrel{(b)}{\le}
\beta_t^2 \|\bv_{t\!-\!1}\! - \!\bg_{t-1}\|^2\! \!\!+ \!\Eb_{\zeta_t}\![2\beta_t^2\|\bar{\w}_t\! -\!\bg_{t}\!\! + \!\bg_{t\!-\!1}\|^2 \!\!\!+\! 2(1\!-\!\beta_t)^2\|\bar{\u}_t \!-\! \bg_{t}\|^2\!] \notag\\
&
\stackrel{(c)}{\le}
\beta_t^2 \|\bv_{t\!-\!1} - \bg_{t\!-\!1}\|^2 + 2\beta_t^2\Eb_{\zeta_t}[\|\bar{\w}_t\|^2] + \frac{2(1-\beta_t)^2\sigma^2}{m} \notag\\
&
\stackrel{(d)}{\le}
\beta_t^2 \|\bv_{t\!-\!1}\! -\! \bg_{t\!-\!1}\|^2 \!+\! \frac{2\beta_t^2}{m}\sum_{i=1}^{m}\Eb_{\zeta_t}[\|\w_{i,t}\|^2] + \frac{2(1-\beta_t)^2\sigma^2}{m} \displaybreak[3] \notag\\
&
\stackrel{(e)}{\le}
\beta_t^2 \|\bv_{t\!-\!1}\! -\! \bg_{t\!-\!1}\|^2\!\! +\! \frac{2\beta_t^2L^2}{m}\sum_{i=1}^{m}\Eb_{\zeta_t}[\|\x_{i,t}\!-\!\x_{i,t\!-\!1}\|^2]\! +\! \frac{2(1\!-\!\beta_t)^2\sigma^2}{m} \notag\\
&
=
\beta_t^2 \|\bv_{t\!-\!1} \!-\! \bg_{t\!-\!1}\|^2\! +\! \frac{2\beta_t^2L^2}{m}\Eb_{\zeta_t}[\|\x_{t}\!-\!\x_{t\!-\!1}\|^2] \!+\! \frac{2(1\!-\!\beta_t)^2\sigma^2}{m},
\end{align}}}
where (a) is because the cross term has the expectation as zero; 
(b) is by $\|\a+\b\|^2 \le 2\|\a\|^2+2\|\b\|^2;$
(c) is by $\Eb[\|\X-\Eb[\X]\|^2] \le \Eb[\|\X\|^2]$ and Assumption 1 (c);
(d) is because of the Jensen's inequality;
(e) is by the $L$-average smoothness.
Thus, taking the full expectation, it holds that 
\begin{align}
\Eb[\|\bv_t - \bg_t \|^2 ]
\le
&
\beta_t^2 \Eb[\|\bv_{t-1} - \bg_{t-1}\|^2] \notag\\
& + \frac{2\beta_t^2L^2}{m}\Eb[\|\x_{t}-\x_{t-1}\|^2] 
+ \frac{2(1-\beta_t)^2\sigma^2}{m} .
\end{align}
\end{proof}

\subsection{Proof for Lemma 2}

\begin{proof}
From the $L$-smoothness of $f,$ we have:
\begin{align}
&
f(\xb_{t\!+\!1}) 
\!\le\!
f(\xb_{t}) \!-\! \eta_t \langle \nabla f(\xb_{t}) , \bv_t \rangle \!+\! \frac{L\eta_t^2}{2}\|\bv_t\|^2 \notag\\
&
\!=\!
f(\xb_{t}) \!-\! \frac{\eta_t}{2}\|\nabla \!f(\xb_{t})\|^2 \!-\! (\frac{\eta_t}{2}  \!-\! \frac{L\eta_t^2}{2})\|\bv_t\|^2 \!+\! \frac{\eta_t}{2}\!\|\bv_t \!-\!\nabla \!f(\xb_{t}) \|^2 \notag\\
&
\!\le\!
f(\xb_{t}) \!-\! \frac{\eta_t}{2}\!\|\nabla f(\xb_{t})\|^2\!-\! (\frac{\eta_t}{2} \!-\! \frac{L\eta_t^2}{2})\|\bv_t\|^2 \notag\\
& +\! \eta_t\|\bv_t \!-\! \bg_t\|^2 \!+\! \eta_t\|\bg_t\!-\!\nabla f(\xb_{t}) \|^2 \notag\\
&
\stackrel{(a)}{\le}
f(\xb_{t}) \!-\! \frac{\eta_t}{2}\|\nabla f(\xb_{t})\|^2 \!-\! (\frac{\eta_t}{2} \! -\! \frac{L\eta_t^2}{2})\|\bv_t\|^2 \notag\\
&+\! \eta_t\|\bv_t\! -\! \bg_t\|^2 \!+\! \frac{\eta_t}{m}\sum_{i=1}^{m}\|\g_{i,t}\!-\!\nabla f_i(\xb_{t}) \|^2 \notag\\
&
\stackrel{(b)}{\le}
f(\xb_{t}) \!-\! \frac{\eta_t}{2}\|\nabla f(\xb_{t})\|^2 \!-\! (\frac{\eta_t}{2}  \!-\! \frac{L\eta_t^2}{2})\|\bv_t\|^2 \notag\\
&+\! \eta_t\|\bv_t \!-\! \bg_t\|^2 \!+\! \frac{L^2\eta_t}{m}\|\x_t\!-\!\1\otimes\xb_{t} \|^2 ,
\end{align}
where $\bg_t = \frac{1}{m}\sum_{i=1}^{m} \g_{i,t}$ and $\g_{i,t} = \nabla f_i(\x_{i,t}),$  (a) is because of the Jensen's inequality and (b) is by the $L$-average smoothness. 
Take the full expectation on the above inequality:
\begin{align}
\Eb[f(\xb_{t\!+\!1})]\!-\!\Eb[f(&\xb_{t})]
\!\le\! 
 -\! \frac{\eta_t}{2}\Eb[\|\nabla f(\xb_{t})\|^2] \!-\! (\frac{\eta_t}{2} \!-\! \frac{L\eta_t^2}{2})\Eb[\|\bv_t\|^2]  \notag\\
 &
 \!+\! \eta_t\Eb[\|\bv_t \!-\! \bg_t\|^2] \!+\! \frac{L^2\eta_t}{m}\Eb[\|\x_t\!-\!\1\!\otimes\!\xb_{t} \|^2 ].
\end{align}
\end{proof}

\subsection{Proof for Lemma 3}

\begin{proof}
First for the iterate $\x_t,$ we have the following contraction:
\begin{align}
\|\tilde{\W}\x_{t} \!-\!\1\otimes\xb_{t} \|^2\! =\! \|\tilde{\W}(\x_{t}\! -\!\1\!\otimes\!\xb_{t}) \|^2 \!\le\! \lambda^2\|\x_{t} \!-\!\1\!\otimes\!\xb_{t}\|^2,
\end{align}
This is because $\x_{t} -\1\otimes\xb_{t}$ is orthogonal to $\1,$ which is the eigenvector corresponding to the largest eigenvalue of $\tilde{\W},$ and $\lambda = \max\{|\lambda_2|,|\lambda_m|\}.$
Recall that $\xb_{t+1} = \xb_{t} - \eta_{t}\bv_{t},$ hence,
\begin{align}\label{Eq: x_contraction}
&\|\x_{t+1}-\1\otimes\xb_{t+1}\|^2
=
\|\tilde{\W}\x_{t} -\eta_t\v_{t}-\1\otimes(\xb_{t} -\eta_{t}\bv_{t})\|^2 \notag\\
&
\le
(1+c_1)\|\tilde{\W}\x_{t} -\1\otimes\xb_{t} \|^2 + (1+\frac{1}{c_1}) \eta_{t}^2\|\v_{t}-\1 \otimes\bv_{t}\|^2 \notag\\
&
\le
(1+c_1)\lambda^2\|\x_{t} -\1\otimes\xb_{t} \|^2 + (1+\frac{1}{c_1}) \eta_{t}^2\|\v_{t}-\1\otimes \bv_{t}\|^2.  
\end{align}
Similarly to (\ref{Eq: x_contraction}), we have:
\begin{align}
&\|\v_{t\!+\!1}\!-\!\1\!\otimes\!\bv_{t\!+\!1}\|^2  \notag\\
\!=&
\|\beta_{t\!+\!1} (\tilde{\W}\v_{t}\! +\! \w_{t\!+\!1}) \!+\! (1\!-\!\beta_t)\u_{t\!+\!1} \!-\! \1 \!\otimes\!\big(\beta_{t\!+\!1} (\bv_{t} \notag\\
&+\! \bar{\w}_{t\!+\!1}) \!+\! (1\!-\!\beta_{t\!+\!1})\bar{\u}_{t\!+\!1}\big)\| \notag\\
\!\le&
(1\!+\!c_1)\beta_{t\!+\!1}^2\lambda^2\|\v_{t} \!-\! \1 \!\otimes\!\bv_{t}\|^2 
\!+\! (1 \!+\! \frac{1}{c_1})\|\beta_{t\!+\!1}(\w_{t\!+\!1}\notag\\
&
\!-\!\1\!\otimes\!\bar{\w}_{t\!+\!1}) \!+\! (1\!-\!\beta_{t\!+\!1})(\u_{t\!+\!1}\!-\!\1\!\otimes\!\bar{\u}_{t\!+\!1})\|^2\notag\displaybreak[4]\\
\!\le&
(1\!+\!c_1)\beta_{t\!+\!1}^2\lambda^2\|\v_{t} \!-\! \1\!\otimes\! \bv_{t}\|^2\!+\! 2(1 \!+\! \frac{1}{c_1})\big(\beta_{t\!+\!1}^2\|\w_{t\!+\!1}\notag\\
&\!-\!\1\!\otimes\!\bar{\w}_{t\!+\!1}\|^2 \!+\! (1\!-\!\beta_{t\!+\!1})^2\|\u_{t\!+\!1}\!-\!\1\!\otimes\!\bar{\u}_{t\!+\!1}\|^2\big)\notag\\
\!\stackrel{(a)}{\le}\!&
(1\!+\!c_1)\beta_{t\!+\!1}^2\lambda^2\|\v_{t} \!-\! \1\!\otimes\! \bv_{t}\|^2 \notag\\
&\!+\! 2(1 \!+\! \frac{1}{c_1})\big(\beta_{t\!+\!1}^2\|\w_{t\!+\!1}\|^2 \!+\! (1\!-\!\beta_{t\!+\!1})^2\|\u_{t\!+\!1}\|^2\big)
\end{align}
where (a) is due to $\|\I-\frac{1}{m}\1\1^\top \|\le 1.$ Lastly, according to the updating equation (9) in main paper, it holds
\begin{align}
&
\|\x_{t+1}-\x_{t}\|^2 
= 
\|\tilde{\W} \x_{t} -\eta_{t} \v_{t} - \x_{t}\|^2 \notag\\
=
&
\|(\tilde{\W} -\I)\x_{t} - \eta_{t}\v_{t}\|^2
\le 2\|(\tilde{\W} -\I)\x_{t} \|^2 + 2\eta_{t}^2 \|\v_{t}\|^2 \notag\\
=
&
2\|(\tilde{\W} -\I)(\x_{t} - \1\otimes\xb_{t}) \|^2 + 2\eta_{t}^2 \|\v_{t}\|^2 \notag\\
\stackrel{(a)}{\le}
&
8\|(\x_{t} - \1\otimes\xb_{t}) \|^2 + 4\eta_{t}^2 \|\v_{t} - \1\otimes\bv_{t}\|^2 + 4\eta_{t}^2m\|\bv_{t}\|^2, 
\end{align}
where (a) is because $\|\tilde{\W} -\I\| \le 2.$
\end{proof}

\subsection{Proof for Lemma 4}

\begin{proof}
First, with $\eta_t = \tau/(\omega + t)^{1/3},$ we have:
\begin{align}\label{Eq: 1/eta_t-1/eta_t-1}
\frac{1}{\eta_t} - \frac{1}{\eta_{t-1}}
& = \frac{1}{\tau}\big((\omega + t)^{\frac{1}{3}} - (\omega + t - 1)^{\frac{1}{3}}\big)\notag\\
& 
\stackrel{(a)}{\le} 
\frac{1}{3\tau} \cdot \frac{1}{(\omega + t - 1)^{\frac{2}{3}}} = \frac{1}{3\tau^3} \eta_{t-1}^2 \notag\\
&
\stackrel{(b)}{\le} 
\frac{1}{3\tau} \cdot \frac{1}{(\frac{\omega}{2} + t)^{\frac{2}{3}}} 
\stackrel{}{\le} 
\frac{1}{3\tau} \cdot \frac{2^{\frac{2}{3}}}{(\omega + t)^{\frac{2}{3}}} \notag\\
&
\stackrel{}{\le} 
\frac{2^{\frac{2}{3}}}{3\tau^3} \cdot \frac{\tau^2}{(\omega + t)^{\frac{2}{3}}} \le \frac{2}{3\tau^3}\eta_t,
\end{align}
where (a) is by $(x+y)^{1/3} - x^{1/3} \le  yx^{-2/3}/3$ and (b) is by $\omega \ge 2.$

Then, we give the following three contractions:

i) for $\Eb[\|\bv_t - \bg_t\|^2],$ we have:
\begin{align}
&
\frac{1}{\eta_t}\Eb[\|\bv_{t\!+\!1}\! -\! \bg_{t\!+\!1}\|^2] \!-\! \frac{1}{\eta_{t\!-\!1}}\Eb[\|\bv_{t}\! -\! \bg_{t}\|^2]\notag\\
&
\!\stackrel{(a)}{\le}\!\!
\Big(\frac{\beta_{t\!+\!1}^2}{\eta_t} \!-\! \frac{1}{\eta_{t\!-\!1}}\!\Big) \Eb[\|\bv_{t} \!-\! \bg_{t}\|^2] \!\!+ \!\!\frac{2\beta_{t\!+\!1}^2L^2}{m\eta_t}\Eb[\|\x_{t\!+\!1}\!-\!\x_{t}\|^2\!] 
\!\!+\! \frac{2(1\!\!-\!\beta_{t\!+\!1})^2\sigma^2}{m\eta_t} \notag\\
&
\!\stackrel{(b)}{\le}\!\!
\Big(\!\frac{1\!-\!\rho \eta_{t}^2}{\eta_t} \!-\! \frac{1}{\eta_{t\!-\!1}}\!\Big) \Eb[\|\bv_{t}\! - \!\bg_{t}\|^2\!] \!+\! \frac{2\beta_{t\!+\!1}^2L^2}{m\eta_t}\Eb[\|\x_{t+1}-\x_{t}\|^2\!] 
\!\!+\! \frac{2\rho^2\sigma^2\eta_t^3}{m} \notag\\
&
\!\stackrel{(c)}{\le}\!\!
\big(\!\frac{2}{3\tau^3} \!-\!\rho \!\big)\eta_{t} \Eb[\|\bv_{t}\! - \!\bg_{t}\|^2] \!+\! \frac{2\beta_{t\!+\!1}^2L^2}{m\eta_t}\Eb[\|\x_{t\!+\!1}\!-\!\x_{t}\|^2] 
\!+\! \frac{2\rho^2\sigma^2\eta_t^3}{m} \notag\\
&
\!\stackrel{(d)}{=}\!\!
-32\eta_{t} L^2\Eb[\|\bv_{t} \!- \!\bg_{t}\|^2] \!\!+\! \frac{2\beta_{t\!+\!1}^2L^2}{m\eta_t}\Eb[\|\x_{t\!+\!1}\!-\!\x_{t}\|^2] 
\!\!+\! \frac{2\rho^2\sigma^2\eta_t^3}{m} 
\end{align}
where (a) is from Lemma \ref{Lemma: Error of v}, (b) is by $\beta_{t+1} = 1 - \rho\eta^2_t < 1,$ (c) is by (\ref{Eq: 1/eta_t-1/eta_t-1}) and (d) is by the setting $\rho = 2/(3\tau^3) + 32 L^2.$

ii) for $\Eb[\|\x_{t}-\1\otimes\xb_{t}\|^2]$, we have:
\begin{align}
&
\frac{1}{\eta_t}\Eb[\|\x_{t\!+\!1}\!-\!\1\!\otimes\!\xb_{t\!+\!1}\|^2] \!-\! \frac{1}{\eta_{t\!-\!1}}\Eb[\|\x_{t}\!-\!\1\!\otimes\!\xb_{t}\|^2] \notag\\
&
\!\stackrel{(a)}{\le}\!\!
\Big(\!\frac{(1\!+\!c_1)\lambda^2}{\eta_t}\!-\!\frac{1}{\eta_{t\!-\!1}}\!\Big)[\|\x_{t}\!-\!\1\!\otimes\!\xb_{t}\|^2] \!+\! (1\!+\!\frac{1}{c_1})\eta_t\Eb[\|\v_{t}\!-\!\1\!\otimes \!\bv_{t}\|^2]\notag\\
&
\!\stackrel{(b)}{\le}\!\!
\Big(\!\frac{(1\!+\!c_1)\lambda^2\!\!-\!1}{\eta_t} \!+\! \frac{2}{3\tau^3}\eta_t\!\Big)[\|\x_{t}\!-\!\1\!\otimes\!\xb_{t}\|^2] \!\!+\! (1\!\!+\!\!\frac{1}{c_1})\eta_t\Eb[\|\v_{t}\!\!-\!\1 \!\otimes\!\bv_{t}\|^2]
\end{align}
where (a) is by Lemma \ref{Lemma: Iterates Contraction} and (b) is by (\ref{Eq: 1/eta_t-1/eta_t-1}).

iii) for $\Eb[\|\v_{t}-\1\otimes\bv_{t}\|^2]$, we have:
\begin{align}
&
\Eb[\|\v_{t+1}-\1\otimes\bv_{t+1}\|^2 - \Eb[\|\v_{t}-\1\otimes\bv_{t}\|^2] \notag\\
&
\stackrel{(a)}{\le}
\big((1+c_1)\beta_{t+1}^2\lambda^2-1\big)\Eb[\|\v_{t} - \1 \otimes\bv_{t}\|^2]\notag\\
&
~~~~~~~~~~~~~~~~~~~~~~~~~~~~~~+ 2(1 + \frac{1}{c_1})\big(\beta_{t+1}^2\Eb[\|\w_{t+1}\|^2] + (1-\beta_{t+1})^2\Eb[\|\u_{t+1}\|^2]\big)\notag\\
&
\stackrel{(b)}{\le}
\big((1+c_1)\beta_{t+1}^2\lambda^2-1\big)\Eb[\|\v_{t} - \1\otimes\bv_{t}\|^2]\notag\\
&
~~~~~~~~~~~~~~~~~~~~~~~~~~~~~~+ 2(1 + \frac{1}{c_1})\big(\beta_{t+1}^2L^2\Eb[\|\x_{t+1}-\x_t\|^2] + mG^2\rho^2\eta_t^4\big)
\end{align}
where (a) is by Lemma \ref{Lemma: Iterates Contraction} and (b) is by $\beta_{t+1} = 1 - \rho\eta^2_t$ and Assumption 1.

Recall the result from Lemma \ref{Lemma: Descend lemma}:
\begin{align}
\Eb[f(\xb_{t\!+\!1})]\!-&\!\Eb[f(\xb_{t})]
\!\le\! 
 \!-\! \frac{\eta_t}{2}\Eb[\|\nabla f(\xb_{t})\|^2] \!-\! (\frac{\eta_t}{2}  \!-\! \frac{L\eta_t^2}{2})\Eb[\|\bv_t\|^2]  \notag\\
 &
 \!+\! \eta_t\Eb[\|\bv_t \!-\! \bg_t\|^2] \!+\! \frac{L^2\eta_t}{m}\Eb[\|\x_t\!-\!\1\!\otimes\!\xb_{t} \|^2 ].
\end{align}
Then, with the result in i), we have:
\begin{align}\label{Eq: protential 1}
&
\Eb[f(\xb_{t\!+\!1})\!+\! \frac{1}{32L^2\eta_t}\|\bg_{t\!+\!1} \!-\! \bv_{t\!+\!1}\|^2]\!-\!\Eb[f(\xb_{t})\!+\! \frac{1}{32L^2\eta_{t\!\!-\!1}}\|\bg_{t} \!-\! \bv_{t}\|^2] \notag\\
\!\le \!&
 -\! \frac{\eta_t}{2}\Eb[\|\nabla f(\xb_{t})\|^2] \!- \!(\frac{\eta_t}{2}  \!-\!\! \frac{L\eta_t^2}{2})\Eb[\|\bv_t\|^2]  
 \!+\! \frac{\beta_{t\!+\!1}^2}{16m\eta_t}\Eb[\|\x_{t\!+\!1}\!-\!\x_{t}\|^2] \notag\\
 &\!+\!\frac{L^2\eta_t}{m}\Eb[\|\x_t\!-\!\1\otimes\xb_{t} \|^2 ]
\!+\! \frac{\rho^2\sigma^2\eta_t^3}{16mL^2} 
\end{align}

Next, with the results in ii) and iii), we have:
\begin{align}\label{Eq: protential 2}
&
\Eb[\frac{1}{\eta_t}\|\x_{t\!+\!1}\!-\!\1\!\otimes\!\xb_{t\!+\!1}\|^2 \!+\! \|\v_{t\!+\!1}\!-\!\1\!\otimes\!\bv_{t\!+\!1}\|^2]\! \notag\\
&- \!\Eb[\frac{1}{\eta_{t\!-\!1}}\|\x_{t}\!-\!\1\!\otimes\!\xb_{t}\|^2 \!+\! \|\v_{t}\!-\!\1\otimes\bv_{t}\|^2] \notag\\
\!\le 
&
\!- \!\Big(\frac{1\!-\!( 1\!+\!c_1)\lambda^2}{\eta_t} \!-\! \frac{2\eta_t}{3\tau^3} \Big)\Eb[\|\x_{t}\!-\!\1\otimes\xb_{t}\|^2]  \notag\\
&
\!-\! \Big(1\!-\! (1\!+\!c_1)\beta_{t\!+\!1}^2\lambda^2 \!-\! (1+\frac{1}{c_1})\eta_t\Big)\Eb[\|\v_{t} \!-\! \1 \otimes\bv_{t}\|^2] \notag\\
&\!+\! 2(1 \!+\! \frac{1}{c_1})\beta_{t\!+\!1}^2L^2\Eb[\|\x_{t\!+\!1}\!-\!\x_t\|^2] \!+\! 2(1 \!+\! \frac{1}{c_1})mG^2\rho^2\eta_t^4.
\end{align}

Thus, for the defined potential function:
\begin{align}
H_t \!=  \Eb[f(\xb_{t})+  &\frac{1}{32L^2\eta_{t\!-\!1}}\|\bg_{t} \!-\! \bv_{t}\|^2\!\notag\\
& +\! \frac{c_0}{m\eta_{t\!-\!1}}\|\x_{t}\!-\!\1\!\otimes\!\xb_{t}\|^2 \!+ \!\frac{c_0}{m}\|\v_{t}\!-\!\1\!\otimes\!\bv_{t}\|^2],
\end{align}
its differential can be calculated as 
\begin{align}
&
H_{t\!+\!1} \!-\!H_t 
\!\le\!
 -\! \frac{\eta_t}{2}\Eb[\|\nabla f(\xb_{t})\|^2] \!-\! (\frac{\eta_t}{2}  \!-\! \frac{L\eta_t^2}{2})\Eb[\|\bv_t\|^2]  \notag\\
 &
 \!+ \!\Big(\!\frac{\beta_{t\!+\!1}^2}{16m\eta_t} \!+\! 2(1 \!+\! \frac{1}{c_1})\frac{c_0\beta_{t\!+\!1}^2L^2}{m}\!\Big)\Eb[\|\x_{t\!+\!1}\!-\!\x_{t}\|^2]  \notag\\
&
\!- \!\Big(\!1\!-\!( 1\!+\!c_1)\lambda^2 \!-\! \frac{2\eta_t^2}{3\tau^3} \!-\! \frac{L^2\eta_t^2}{c_0} \!\Big)\!\times\!\frac{c_0}{m\eta_t}[\|\x_{t}\!-\!\1\!\otimes\!\xb_{t}\|^2]  \notag\\
&
\!-\! \Big(\!1\!-\! (1\!+\!c_1)\beta_{t\!+\!1}^2\lambda^2\! -\! (1\!+\!\frac{1}{c_1})\eta_t\!\Big)\!\times\!\frac{c_0}{m}\Eb[\|\v_{t} \!- \!\1 \!\otimes\!\bv_{t}\|^2] \notag\\
&
\!+\! \frac{\rho^2\sigma^2\eta_t^3}{16mL^2}
\!+\! 2(1 \! +\! \frac{1}{c_1})c_0G^2\rho^2\eta_t^4\notag\\
\!\stackrel{(a)}{\le}\!
&
 - \!\frac{\eta_t}{2}\Eb[\|\nabla f(\xb_{t})\|^2] \!+\! \frac{\rho^2\sigma^2\eta_t^3}{16mL^2}
\!+\! 2(1 \!+\! \frac{1}{c_1})c_0G^2\rho^2\eta_t^4 \notag\displaybreak[4]\\
&
\!-\! \Big(\!1\!-\!( 1\!+\!c_1)\lambda^2 \!- \!\frac{1}{2c_0} \!-\! 16(1 \!+\! \frac{1}{c_1})L^2\eta_t \!- \! \frac{2\eta_t^2}{3\tau^3} \notag\\
&
\!-\! \frac{L^2\eta_t^2}{c_0} \! \Big)\!\times\!\frac{c_0}{m\eta_t}\Eb[\|\x_{t}\!-\!\1\!\otimes\!\xb_{t}\|^2]  \notag\\
&
\!- \!\Big(\!1\!-\! (1\!+\!c_1)\lambda^2 \!-\! (1\!+\!\frac{1}{c_1})\eta_t \!-\! \frac{\eta_t}{4c_0} \notag\\
&\!-\! 8(1 \!+\! \frac{1}{c_1})L^2\eta_{t}^2 \!\Big)\!\times\!\frac{c_0}{m}\Eb[\|\v_{t} \!-\! \1\!\otimes\! \bv_{t}\|^2] \notag\\
 &
 \!-\! \Big(\!1  \!-\! 2L\eta_t \!-\! 32(1 \!+\! \frac{1}{c_1})c_0L^2\eta_t \!\Big)\!\times \!\frac{\eta_t}{4}\Eb[\|\bv_t\|^2].
\end{align}
where (a) follows from plugging the result for $\|\x_{t+1}-\x_t\|^2$ from Lemma \ref{Lemma: Iterates Contraction} and $\beta_{t+1} < 1.$

\end{proof}

\subsection{Proof for Theorem 1}

\begin{proof}
From Lemma \ref{Lemma: Potential Func}, we have:
\begin{align}
&\sum_{t=0}^{T} \frac{\eta_t}{2}\Eb[\|\nabla f(\xb_{t})\|^2] 
\!\le\!
H_0\!-\! H_{T\!+\!1}  \!+\! \sum_{t=0}^{T}\frac{\rho^2\sigma^2\eta_t^3}{16mL^2}\notag\\
&
\!+ \!\sum_{t=0}^{T} 2(1 \!+ \!\frac{1}{c_1})c_0G^2\rho^2\eta_t^4 
\!-\! \sum_{t=0}^{T} \frac{c_0C_1}{m\eta_t}\Eb[\|\x_{t}\!-\!\1\!\otimes\!\xb_{t}\|^2] \notag\\
&
\!-\! \sum_{t=0}^{T} \frac{c_0C_2}{m}\Eb[\|\v_{t} \!- \!\1\! \otimes\!\bv_{t}\|^2] 
 \!-\! \sum_{t=0}^{T} \frac{C_3\eta_t}{4}\Eb[\|\bv_t\|^2]. 
\end{align}
Note with $\eta_t = \tau / (\omega + t)^{1/3}$ and $\tau \ge 2,$ thus
\begin{align}
&\sum_{t=0}^{T}\eta_t^3 = \sum_{t=0}^{T}\frac{\tau}{\omega + t} \le \int_{-1}^{T-1} \frac{\tau}{\omega + t} dt\le \tau\ln(\omega + T - 1)\\
&\sum_{t=0}^{T}\eta_t^4 = \sum_{t=0}^{T}\Big(\frac{\tau}{\omega + t}\Big)^{\frac{4}{3}} \le \int_{-1}^{T-1} \Big(\frac{\tau}{\omega + t}\Big)^{\frac{4}{3}} dt \le \frac{3\tau^{4/3}}{(\omega - 1)^{1/3}}.
\end{align}
Hence, since $\eta_t$ is decreasing, we have:
\begin{align}
&
\frac{\eta_T}{2}\!\sum_{t=0}^{T} \Eb[\|\nabla f(\xb_{t})\|^2]\!+ \!\frac{1}{m}\Eb[\|\x_{t}\!-\!\1\!\otimes\!\xb_{t}\|^2]   \notag\\
&
\le
\sum_{t=0}^{T}\! \frac{\eta_t}{2}\Eb[\|\nabla f(\xb_{t})\|^2]\!+\! \frac{\eta_t}{2m}\Eb[\|\x_{t}\!-\!\1\!\otimes\!\xb_{t}\|^2] \notag\\
&
\!\le\!
H_0\!-\! H_{T\!+\!1}  
\!+\! \frac{\tau\rho^2\sigma^2\ln(\omega\!+\!T\!-\!1)}{16mL^2} \notag\\
&
\!+\! 6(1 \!+\! \frac{1}{c_1})c_0G^2\rho^2\frac{\tau^{4/3}}{(\omega\!-\!1)^{1/3}}
\!-\! \sum_{t=0}^{T} \frac{2c_0C_1\! - \!\eta_t^2}{2m\eta_t}\Eb[\|\x_{t}\!-\!\1\!\otimes\!\xb_{t}\|^2]  \notag\\
&
\!- \!\sum_{t=0}^{T} \frac{c_0C_2}{m}\Eb[\|\v_{t} \!- \!\1 \!\otimes\!\bv_{t}\|^2] 
 \!- \!\sum_{t=0}^{T} \frac{C_3\eta_t}{4}\Eb[\|\bv_t\|^2]. 
\end{align}

Now, we show that by properly choosing $\eta_t,$ $c_1,$ and $c_0$, the coefficients $C_1 - \eta^2/2c_0,$ $C_2$ and $C_3$ can be non-negative. Recall that:
\begin{align}
C_1 & \!= \!1\!-\!( 1\!+\!c_1)\lambda^2 \!- \!\frac{1}{2c_0} \!- \!16(1 \!+\! \frac{1}{c_1})L^2\eta_t \!- \! \Big(\frac{2}{3\tau^3} \!+\! \frac{L^2}{c_0}\Big)\eta_t^2, \label{Eq: C_1}\\
C_2 &\! =\! 1\!-\! (1\!+\!c_1)\lambda^2 \!- \!(1\!+\!\frac{1}{c_1})\eta_t \!-\! \frac{\eta_t}{4c_0} \!- \!8(1\!+\! \frac{1}{c_1})L^2\eta_{t}^2, \label{Eq: C_2}\\
C_3 & \!=\! 1 \! - \!2L\eta_t \!-\! 32(1 \!+\! \frac{1}{c_1})c_0L^2\eta_t. \label{Eq: C_3}
\end{align}

In order to have $C_3 \ge 0,$ we have:
\begin{align}\label{Eq: C3_2}
\eta_t \le 1/\Big(2L + 32(1 + \frac{1}{c_1})c_0L^2\Big):=k_1.
\end{align}
With (\ref{Eq: C3_2}), it follows that: 
\begin{align}
C_2 & \ge 1- (1+c_1)\lambda^2 - (1+\frac{1}{c_1})\eta_t - \frac{\eta_t}{2c_0}. 
\end{align}
Thus, $C_2\ge 0$ if we set 
\begin{align}\label{Eq: C2_2}
\eta_t \le \Big(1- (1+c_1)\lambda^2\Big) / \Big(1+\frac{1}{c_1} + \frac{1}{2c_0}\Big):=k_2.
\end{align}
For $C_1 - \eta^2/2c_0,$ it follows from (\ref{Eq: C3_2}) that:
\begin{align}
C_1 - \frac{\eta^2}{2c_0} & \ge 1-( 1+c_1)\lambda^2 - \frac{1}{c_0} -  \Big(\frac{2}{3\tau^3} + \frac{2L^2+1}{2c_0}\Big)\eta_t^2.
\end{align}
By choosing 
\begin{align}
\eta_t &\le \sqrt{\Big(1-( 1+c_1)\lambda^2 - \frac{1}{c_0}\Big)/\Big(\frac{2}{3\tau^3} + \frac{2L^2+1}{2c_0}\Big)}:=k_3, \\
\text{and }~0 &< 1-( 1+c_1)\lambda^2 - \frac{3}{4c_0},
\end{align}
we have $C_1 - \eta^2/2c_0 \ge 0.$
To summarize, we need to set $\eta_t \le \min\{k_1,k_2,k_3\}.$
Since $\eta_t$ is decreasing and $\eta_0 = \tau/\omega^{1/3},$ it implies that $\omega \ge (\tau/\min\{k_1,k_2,k_3\})^3.$

With the above parameter setting, we have:
\begin{align}
&\frac{\eta_T}{2}\sum_{t=0}^{T} \Eb[\|\nabla f(\xb_{t})\|^2]+ \frac{1}{m}\Eb[\|\x_{t}-\1\otimes\xb_{t}\|^2] 
\le
H_0- H_{T+1}   \notag\\
&
+ \frac{\tau\rho^2\sigma^2\ln(\omega+T-1)}{16mL^2}
+ 6(1 + \frac{1}{c_1})c_0G^2\rho^2\frac{\tau^{4/3}}{(\omega-1)^{1/3}}.
\end{align}
Multiplying both side of the above inequality by ${2}/{\eta_T(T+1)},$ we have:
\begin{align}
&
\frac{1}{T\!+\!1}\!\sum_{t=0}^{T} \Eb[\|\nabla f(\xb_{t})\|^2]\!+ \!\frac{1}{m}\Eb[\|\x_{t}\!-\!\1\!\otimes\!\xb_{t}\|^2] \notag\\
&
\!\le\!\frac{2(H_0\!-\! H_{T\!+\!1})}{\eta_T(T\!+\!1)}  
\!+\! \frac{\tau\rho^2\sigma^2\ln(\omega\!+\!T\!-\!1)}{8mL^2\eta_T(T\!+\!1)}
\!+ \!\frac{12(1 \!+ \!\frac{1}{c_1})c_0\tau^{4/3}G^2\rho^2}{(\omega\!-\!1)^{1/3}\eta_{T}(T\!+\!1)}.
\end{align}

Note that
 \begin{align}
H_{0} 
&
\!=\!\Eb[f(\xb_{0})\!+\! \frac{1}{32L^2\eta_{-1}}\|\bg_{0}\! - \!\bv_{0}\|^2 \notag\\
&\!+ \!\frac{c_0}{m\eta_{-1}}\|\x_{0}\!-\!\1\!\otimes\!\xb_{0}\|^2 \!+\! \frac{c_0}{m}\|\v_{0}\!-\!\1\!\otimes\!\bv_{0}\|^2]\ \notag\\
& 
\!\stackrel{(a)}{=}\!
\Eb[f(\xb_{0})\!+\! \frac{1}{32L^2\eta_{-1}}\|\bg_{0} \!- \!\bv_{0}\|^2 \!+\! \frac{c_0}{m}\|\v_{0}\!-\!\1\!\otimes\!\bv_{0}\|^2]\ \notag\\
& 
\!\stackrel{(b)}{\le}\!
\Eb[f(\xb_{0})\!+ \!\frac{\sigma^2}{32mL^2\eta_{-1}} \!+ \!\frac{c_0}{m}\|\v_{0}\!-\!\1\!\otimes\!\bv_{0}\|^2] \notag\\
H_{T\!+\!1} 
&
\!\ge\! \Eb[f(\xb_{T\!+\!1}) \!+\!\frac{c_0}{m\eta_T} \|\x_{T\!+\!1}\!-\!\1\xb_{T\!+\!1}\|^2 \!+\! \frac{c_0}{m}\|\v_{T\!+\!1}\!-\!\1\!\otimes\!\bv_{T\!+\!1}\|^2]\notag\\
&
 \ge f(\xb^*) \notag,
 \end{align}
 where (a) is by $\x_{i,0} = \x_0$ from line 1 in Algorithm 1 and (b) is by Assumption 1.

Hence, it follows that 
\begin{align}
&\frac{1}{T\!+\!1}\sum_{t=0}^{T} \Eb[\|\nabla f(\xb_{t})\|^2]\!+\! \frac{1}{m}\Eb[\|\x_{t}\!-\!\1\!\otimes\!\xb_{t}\|^2] \notag\\
&
\!\le\!
\frac{2(f(\xb_{0})  \!- \!f(\xb^*))}{\eta_T{(T\!+\!1)}}  
\!+\! \frac{2c_0\Eb[\|\v_{0}\!-\!\1\!\otimes\!\bv_{0}\|^2]}{m\eta_T{(T\!+\!1)}} \notag\\
&
\!+ \!\frac{(\omega \!- \!1)\sigma^2}{16mL^2\tau\eta_T{(T\!+\!1)}} 
\!+\! \frac{\tau\rho^2\sigma^2\ln(\omega\!+T\!-\!1)}{8mL^2\eta_T(T\!+\!1)}\notag\\
&
\!+\! 12(1 \!+\!\frac{1}{c_1})c_0\frac{\tau^{4/3}G^2\rho^2}{(\omega \!-\! 1)^{1/3}\eta_{T}(T\!+\!1)}.
\end{align}
Since $\eta_T= \tau/(\omega+T)^{1/3}$, we have:
\begin{align}
&
\min_{t\in[T]}\! \Eb[\|\nabla f(\xb_{t})\|^2]\!+\! \frac{1}{m}\Eb[\|\x_{t}\!-\!\1\!\otimes\!\xb_{t}\|^2] \notag\\
&
\!\le\!
\frac{2(f(\xb_{0})  \!- \!f(\xb^*))}{\tau{(T\!+\!1)}^{2/3}}  
\!+\! \frac{2c_0\Eb[\|\v_{0}\!-\!\1\!\otimes\!\bv_{0}\|^2]}{m\tau{(T\!+\!1)}^{2/3}} \notag\\
&
\!+ \!\frac{(\omega\! - \!1)\sigma^2}{16mL^2\tau^2{(T\!+\!1)}^{2/3}} 
\!+\! \frac{\rho^2\sigma^2\ln(\omega\!+\!T\!-\!1)}{8mL^2(T\!+\!1)^{2/3}}\notag\\
&
\!+\! \frac{12(1\! +\! \frac{1}{c_1})c_0\tau^{1/3}G^2\rho^2}{(\omega \!-\! 1)^{1/3}(T\!+\!1)^{2/3}}\! +\!O\Big(\frac{c_3\omega}{\tau T^{5/3}}\Big).
\end{align}
where the $O$-notation is from $(\omega+T)^{1/3} - (T+1)^{1/3} \le (\omega-1)(T+1)^{-2/3}/3$ and $c_3 = \max\{1,(\omega-1)/(m\tau^2), \tau^{4/3}/\omega^{1/3}, \tau \ln(\omega +T -1)/m\}.$

\end{proof}

\subsection{Proof for Corollary 2}

\begin{proof}
First, note that $\omega \ge \max\{2,\tau^3/\min\{k_1^3,k_2^3,k_3^3\}\}$ holds with $\tau = O(m^{1/3})$ and $\omega = O(m^{4/3}).$
Plugging these parameters into Theorem 1 yields: 
\begin{align}
&\min_{t\in[T]}\! \Eb[\|\nabla f(\xb_{t})\|^2]\!+\! \frac{1}{m}\Eb[\|\x_{t}\!-\!\1\!\otimes\!\xb_{t}\|^2] \notag\\
&
\!\le\!
O\Big(\frac{2(f(\xb_{0}) \! - \!f(\xb^*))}{m^{1/3}{(T\!+\!1)}^{2/3}}  
\!+ \!\frac{2c_0\Eb[\|\v_{0}\!-\!\1\!\otimes\!\bv_{0}\|^2]}{m^{4/3}{(T\!+\!1)}^{2/3}}
\!+\! \frac{\sigma^2}{16L^2m^{1/3}{(T\!+\!1)}^{2/3}}  \notag\\
&
\!+\! \frac{\rho^2\sigma^2\ln(m^{4/3}\!+\!T)}{8mL^2(T\!+\!1)^{2/3}}
\!+\! \frac{12(1 + \frac{1}{c_1})c_0G^2\rho^2}{m^{1/3}(T\!+\!1)^{2/3}} \!+\! \frac{c_3m}{T^{5/3}}\Big).
\end{align}
With $T \gg m^{4/3},$ we have:
\begin{align}
&\min_{t\in[T]}\! \Eb[\|\nabla f(\xb_{t})\|^2]\!+\! \frac{1}{m}\Eb[\|\x_{t}\!-\!\1\!\otimes\!\xb_{t}\|^2] 
\notag\\
&
\!\le\!
O\Big(\frac{2(f(\xb_{0})  \!-\! f(\xb^*))}{m^{1/3}{(T\!+\!1)}^{2/3}}  
\!+ \!\frac{2c_0\Eb[\|\v_{0}\!-\!\1\!\otimes\!\bv_{0}\|^2]}{m^{4/3}{(T\!+\!1)}^{2/3}} \notag\\
&
\!+\! \frac{\sigma^2}{16L^2m^{1/3}{(T\!+\!1)}^{2/3}} 
\!+\! \frac{\rho^2\sigma^2\ln T}{8mL^2(T\!+\!1)^{2/3}} \notag\\
&
+ \frac{12(1 \!+\! \frac{1}{c_1})c_0G^2\rho^2}{m^{1/3}(T\!+\!1)^{2/3}} \!+\! \frac{c_3}{m^{1/3}T^{2/3}}\Big),
\end{align}
where $c_3 = \max\{O(1), O(1/m^{1/3}), \ln(m^{4/3} +T)/m^{2/3}\}.$
The above result implies that the convergence rate is $\tilde{O}(m^{-1/3}T^{-2/3}).$
Thus, to achieve an $\epsilon^2$-stationary solution, the total communication rounds needed are
 $T = \tilde{O}(m^{-1/2}\epsilon^{-3})$, and 
the total samples needed are $mT = \tilde{O}(m^{1/2}\epsilon^{-3}).$
\end{proof}

\end{document}